%% file: root.tex
\documentclass[letterpaper, 10 pt, conference]{ieeeconf}  % Comment this line
\IEEEoverridecommandlockouts                              % This command is only needed if
\small
\title{Learning Human Rewards by Inferring Their Latent Intelligence Levels in Multi-Agent Games: A Theory-of-Mind Approach with Application to Driving Data}
\normalsize

\author{Ran Tian, Masayoshi Tomizuka, and Liting Sun
  \thanks{
         Ran Tian, Masayoshi Tomizuka, and Liting Sun are with University of California, Berkeley. ({\tt\small \{rantian, tomizuka, litingsun\}@berkeley.edu}).
     }
}
\usepackage{tabularx}
\usepackage{float}
\usepackage[caption=false]{subfig}
\usepackage{epsfig}
\usepackage{epstopdf}
\usepackage{bbm}
\usepackage{mathtools,xparse}

\usepackage{epsfig} % for postscript graphics files
\usepackage{amsmath} % assumes amsmath package installed
\usepackage{amssymb}  % assumes amsmath package installed
\usepackage{bm}
\usepackage{epstopdf}
\usepackage{color}
\usepackage{etoolbox}
\DeclareMathOperator*{\argmax}{arg\,max}

\usepackage{comment}
\usepackage{diagbox}
\usepackage[ruled, lined, longend, linesnumbered]{algorithm2e}
\usepackage{booktabs,multirow}
\usepackage{bigstrut}
\usepackage{cite}

\usepackage{lipsum}
\usepackage{color}
\usepackage[colorinlistoftodos]{todonotes}
\usepackage{hyperref}
\usepackage{balance}

\usepackage{amsfonts}
\usepackage{mathtools}
\usepackage{graphicx}
\usepackage[capitalise]{cleveref}
\usepackage{hyperref}
\begin{document}
\maketitle

\begin{abstract}
Reward function, as an incentive representation that recognizes humans' agency and rationalizes humans' actions, is particularly appealing for modeling human behavior in human-robot interaction. Inverse Reinforcement Learning is an effective way to retrieve reward functions from demonstrations. However, it has always been challenging when applying it to multi-agent settings since the mutual influence between agents has to be appropriately modeled. To tackle this challenge, previous work either exploits equilibrium solution concepts by assuming humans as perfectly rational optimizers with unbounded intelligence or pre-assigns humans' interaction strategies \textit{a priori}. In this work, we advocate that humans are bounded rational and have different intelligence levels when reasoning about others' decision-making process, and such an inherent and latent characteristic should be accounted for in reward learning algorithms. Hence, we exploit such insights from Theory-of-Mind and propose a new multi-agent Inverse Reinforcement Learning framework that reasons about humans' latent intelligence levels during learning. We validate our approach in both zero-sum and general-sum games with synthetic agents and illustrate a practical application to learning human drivers' reward functions from real driving data. We compare our approach with two baseline algorithms. The results show that by reasoning about humans' latent intelligence levels, the proposed approach has more flexibility and capability to retrieve reward functions that explain humans' driving behaviors better. 
\end{abstract}

\input{intro}
\input{related_works}

\input{problem_formulation}

\input{cognition_modeling}
\input{IRL_method}

\input{result}

\vspace{-0.2cm}
\section{Discussion}
\noindent
\textbf{Summary.} In this work, we advocated that humans have different levels of sophistication in reasoning about others' behaviors during interactions, and such an aspect should be accounted for during reward function learning. We exploited insights from Theory-of-Mind and proposed a new Maximum Entropy-based multi-agent Inverse Reinforcement Learning framework that reasons about humans' latent intelligence levels during learning. We validated our approach in both zero-sum and general-sum games with ground truth humans, then illustrated a practical application of our approach to learning human drivers' reward functions from real driving data. We compared our approach with the baseline algorithms and provided a detailed experiment analysis. We found that reasoning about humans' levels of intelligence provides more flexibility during learning and helps explain and reconstruct human driving behaviors better. 

\noindent
\textbf{Limitations and future works.} We view our work as a first step into incorporating Theory-of-Mind based bounded intelligence reasoning into multi-agent IRL. One of our approach's limitations is the ability to treat continuous states and actions. Fortunately, the procedure for computing $\pi^{i,k}$ can be realized in an iterative deep $Q$-learning fashion \cite{BoutonRL} and plugged into \cref{alg: RA-MAIRL} seamlessly. Another limitation is related to the assumption about humans' constant intelligence levels during interactions. Such an assumption is commonly used in one-shot games. However, by inspecting traffic data, we noticed that there might be a short period at the beginning of an interaction during which the agents compete for higher levels of intelligence in order to dominate the interaction. We evasion that incorporating such a dynamic attribute of cognitive states into reward learning could potentially gain more flexibility during learning.

%\begin{IEEEkeywords}

%\end{IEEEkeywords}

% \fi
% creates the second title. It wipassivekyll be ignored for other modes.
\IEEEpeerreviewmaketitle

\bibliographystyle{IEEEtran}

%%\bibliography{ref} C:\Users\hp\Documents
\balance
%\small
\bibliography{Ref}

\end{document}

%% file: intro.tex
\section{Introduction}\label{sec: intro}

% Learning reward functions are important
Our society is rapidly advancing towards robots that collaborate with and assist humans in daily tasks. To assure safety and efficiency, such robots need to understand and predict human behaviors. Modeling humans as reward-driven agents is particularly appealing since reward functions recognize humans' agency and rationalize their actions. Inverse Reinforcement Learning (IRL) has been proved to be an effective approach to learning reward functions from human demonstrations. However, most previous IRL work is restricted to single-agent settings \cite{russell1998learning, ng2000algorithms, abbeel2004apprenticeship, ziebart2008maximum, finn2016connection}.

\begin{figure}[t]
\begin{center}
\begin{picture}(300, 105)
%%%%%%%%%%%%%%%%%%%%%%%%%%%%%%
\put( -10,  0){\epsfig{file=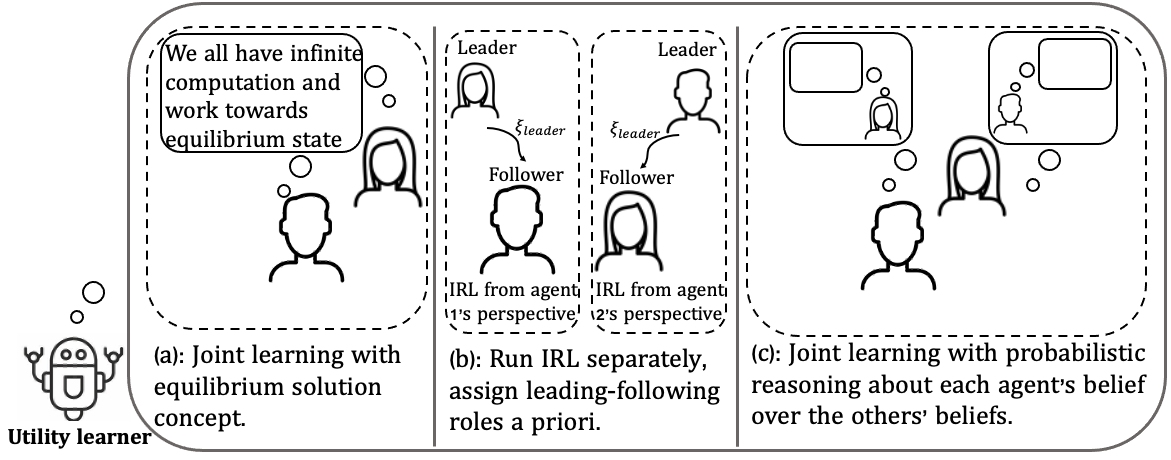,width = 1.05\linewidth, trim=0.0cm 0.0cm 0cm 0cm,clip}}  %%%
\
\end{picture}
\end{center}
\vspace{-0.3cm}
\caption{Different IRL formulations in multi-agent interactions, (a): IRL approaches that exploits equilibrium solution concepts which assume agents have infinite intelligence levels. (b): IRL approaches that assign leader-follower roles \textit{a priori}. (c): our approach that reasons about agents' intelligence levels during reward learning.}
\label{fig: method compare}
\vspace{-0.5cm}
\end{figure}

% Problems and challenges with current approaches in multi-agent IRL
Learning reward functions in multi-agent settings is challenging, as it requires an interaction model that characterizes the mutual influence or the closed-loop dynamics among agents. Theory-of-Mind \cite{goldman2012theory} is a mechanism that explicitly models humans' beliefs over other humans' decision-making process. Based on Theory-of-Mind, when a human interacts with his/her opponent, his/her cognitive reasoning process is nested and can be structured in a recursive fashion: ``I believe that you believe that I believe..." We refer to the depth of such recursion as a human's \textit{intelligence level}, and it characterizes a human's cognitive reasoning ability. Previous work on multi-agent IRL has leveraged equilibrium solution concepts of Markov Games to model interactions \cite{yu2019multi, GruverMulti} as shown in \cref{fig: method compare}(a). Equilibrium solution-based approaches assume that humans have unlimited computation and infinite intelligence levels when making decisions. However, in real life, humans rarely pursue equilibrium solutions \cite{coricelli2009neural}. Hence, to account for such facts, non-equilibrium solution-based approaches have also been explored by many researchers in IRL algorithms. For instance, \cite{sadigh2017active, biyik2018batch, sun_probabilistic_2018, suncourteous} exploited a leader-follower game to account for humans' bounded intelligence in two-agent IRL settings. Instead of jointly learning agents' reward functions, these approaches run IRL from each agent's perspective with pre-assigned leader/follower roles (refer to \cref{sec: related works} for more details).

% What we did and key insight
\textit{In this work, we advocate that humans have latent intelligence levels that are naturally bounded and heterogeneous, and such an aspect should be accounted for in multi-agent IRL formulations.} Specifically, we exploit insights from Theory-of-Mind to explicitly model humans' interactions under bounded intelligence, and propose a multi-agent IRL formulation that reasons about humans' heterogeneous intelligence levels during learning. Our key insight is: \textit{reasoning about humans' hidden intelligence levels (a type of latent cognitive states) adds more flexibility to multi-agent IRL algorithms and helps learn reward functions that explain and reconstruct human behaviors better.}

Overall, we make the following contributions towards multi-agent IRL:

\noindent
\textbf{Developing a framework for joint reward functions learning in multi-agent settings without assumptions about perfect rationality or leader-follower roles.} We exploit insights from Theory-of-Mind to account for humans' bounded intelligence and extend the Maximum Entropy Inverse Reinforcement Learning (MaxEnt IRL) framework in multi-agent settings with explicit reasoning about humans' latent intelligence levels during learning. We validate our approach in both zero-sum and general-sum games with synthetic agents, then apply our approach to learn human driver reward functions from real-world driving data.

\noindent
\textbf{Analyzing the advantages of reasoning about humans' intelligence levels in multi-agent IRL with real driving data.} We conduct, by our knowledge, the first quantitative comparison between two types of IRL algorithms using real driving data: the IRL algorithm that pre-assigns humans' leader-follower roles during learning \cite{sadigh2017active, biyik2018batch, sun_probabilistic_2018, suncourteous} and our approach that leverages the cognitive reasoning structure to reason about humans' latent intelligence levels. We show that our approach provides more flexibility during learning and helps explain and reconstruct real human drivers' behaviors better.

%% file: related_works.tex
\section{Related works}\label{sec: related works}

\noindent
\textbf{Opponent modeling.} Our work is an instance of opponent modeling. Traditional approaches in opponent modeling often model opponents' behaviors based on past experience. In particular, \cite{sadigh2016information, albrecht2017special, fridovich2020confidence} utilized type-based reasoning approaches that assume an opponent agent is one of known types and identify opponents' types online based on past interactions. Other work also modeled opponents in terms of leader-follower relationship: ego agent models its opponents’ responsive actions as a probability distribution conditioned on its own action \cite{sadigh2016planning, fisac2019}. Insights from Theory-of-Mind have been exploited to not only model opponents' conditional distribution of actions but also model opponents' beliefs over other agents' beliefs, such nested beliefs (``I  believe  that  you  believe  that  I believe...") can be explicitly modeled by approaches based on recursive reasoning \cite{baker2011bayesian, li2017game, de2017negotiating, wen2019probabilistic, tian2020game}. Our approach to modeling opponents is based on the quantal level-$k$ model that belongs to the recursive reasoning paradigm. In contrast to \cite{wen2019probabilistic}, we do not assume all agents perform depth-$1$ recursions but recognize the heterogeneity in agents' recursive depths. In contrast to \cite{tian2020game}, we explicitly account for opponents' sub-optimal behaviors.

\noindent
\textbf{Learning human driver reward functions.} Many previous works in learning  human drivers' reward functions are extensions of MaxEnt IRL \cite{ziebart2008maximum} and are restricted to single-agent settings \cite{Sunexpress, wu2020efficient, memarian2020active}. Recently, the concept of quantal best response equilibrium (QRE) is exploited to extend MaxEnt IRL to multi-agent games. Equilibrium solution concepts assume all agents have infinite intelligence levels and have common knowledge of this. However, mounting evidence suggests that human behaviors often deviate from equilibrium behaviors in systematically biased ways due to their compromised / bounded intelligence \cite{goeree2001ten, crawford2007level, coricelli2009neural}. Other approaches have extended MaxEnt IRL in two-agent settings by utilizing a pre-assigned leader-follower relationship to model interactions \cite{sadigh2017active, biyik2018batch, sun_probabilistic_2018, suncourteous}. Consequently, instead of jointly learning reward functions, they have to conduct MaxEnt IRL from each agent's perspective separately. Moreover, in each separate IRL formulation, the interaction is simplified as an open-loop leader-follower game, where the opponent's ground-truth trajectory is assumed to be accessible by the ego agent (\cref{fig: method compare} (b)).
%Therefore, the more accommodations that the ego agent's trajectory brings to the known opponent's trajectory, the more likely that the current roll-out ego agent's reward function is the true reward function, making the ego agent a follower to its opponent.
Therefore, no real interaction has been accounted for in the reward functions learning process.
Furthermore, assigning the leader-follower relationship \textit{a priori} requires careful selection for demonstrations because demonstrations that don't align with the role assignment may lead to biased learning, i.e., incorrect human driver reward functions might be retrieved as demonstrated in \ref{sec:experiment_results}. 

\noindent
\textbf{Comparison between our approach and previous work.} Our approach is different from the approaches in \cite{yu2019multi,GruverMulti} since we relaxed the assumptions about the equilibrium solution concept by leveraging recursive reasoning and recognizing humans' bounded intelligence during learning. Our work is also distinguished from \cite{tianaaai} by extending the learning algorithm to multi-agent settings and illustrating a practical application to learning human driver reward functions from real traffic data. In contrast to \cite{sadigh2017active, biyik2018batch, sun_probabilistic_2018, suncourteous}, our approach jointly learns human drivers' reward functions in multi-agent games without assumptions about agents' roles (\cref{fig: method compare} (c)).

%We also provide detailed experiment analysis and comparison with widely-used IRL algorithms.

%which is a strong assumption and mounting evidence shows that humans rarely follow equilibrium strategies \cite{keynes2018general}.

%In addition, experimental studies suggest that human behaviors often deviate from equilibrium behaviors in systematically biased ways even in simple games due to their compromised / bounded cognition \cite{goeree2001ten, crawford2007level, coricelli2009neural}. 

%Our previous work proposed an IRL framework that exploits non-equilibrium solution concept to model interactions \cite{tianaaai}, but the formulation is for two-agent settings and validated in low-dimension grid-world games. 

%% file: problem_formulation.tex
%\vspace{-0.2cm}
\section{Problem formulation}\label{sec: problem formulation}
\noindent
\textbf{Multi-agent interaction formalization.} We model the interactions between $n$ agents ($n \in\mathbb{N}^+$) as a stochastic game defined by tuple $\langle\mathcal{S}, \mathcal{A}_1,{\dots},\mathcal{A}_n, \mathcal{R}^1,{\dots},\mathcal{R}^n, f, \gamma\rangle$, where $\mathcal{S}$ denotes the finite state space, $\mathcal{A}_i$ denotes the finite action space of agent $i$ ($i{\in}\mathcal{P} {=} \{1,{\dots},n\}$), $\mathcal{R}^i:\mathcal{S}{\rightarrow} \mathbb{R}$ denotes the reward function of agent $i$, $f: \mathcal{S} {\times} \mathcal{A}_1{\times} {\dots} \mathcal{A}_n{\rightarrow} \mathcal{S}$ denotes the open-loop dynamics of the game, and $\gamma$ is the discount factor. For agent $i$, we let $\pi_{i}: \mathcal{S}{\rightarrow}\mathcal{A}_i$ denote its deterministic policy. At each discrete time step $t$, agent $i$ aims to maximize its expected total reward in state $s_t{\in}\mathcal{S}$ {:} $\pi^*_i {=} \argmax_{\pi} V^{\pi}_i(s_t)$, where $ V_{i}^{\pi}(s_t) = \mathbb{E}_{a^{-i}_{t:\infty}} \left[\sum_{\tau=0}^{\infty} \gamma^\tau \mathcal{R}^i(s_{t+\tau}) \big| \pi, f \right]$ is the value function representing agent $i$'s expected return starting from $s_t$, subject to its policy and the dynamics function. The expectation is taken with respect to the possible actions from other agents in the game, denoted by $a^{-i}_t {=} (a^j_t)$ with $j {\in} \mathcal{P}{\setminus}\{i\}$. Different reasoning strategies of the other agents will yield significantly different distribution for $a^{-i}_t$ and impact the closed-loop dynamics of the game.

\noindent
\textbf{Reward learning as an optimization problem.} We consider a reward-free stochastic game. We assume that agent $i$'s reward function is parameterized by a collection parameter $\omega_i$ and $\mathcal{R}^i_{\omega_i}$ is differentiable with respect to $\omega_i$. Such a reward function representation could be a neural network or a linear combination of a given set of features. Suppose we are given a demonstration set $\mathcal{D}$ that contains multiple groups of interaction trajectories among $n$ agents. Our objective is to jointly learn the reward functions for all agents that best rationalize the demonstrations. Namely, we want to find an optimal $\bar{\omega}=(\omega_1,\dots,\omega_n)$ that maximizes the likelihood of observed demonstrations. We optimize for $\bar{\omega}$ in the following maximum likelihood estimation problem:

\small
\begin{align}\label{equ: maximum likelihood}
	\max_{\bar{\omega}} {\sum}_{\xi\in \mathcal{D}} \hspace{-0.3cm}\log \mathbb{P}\left(\xi | \bar{\omega}\right)= \max_{\bar{\omega}} {\sum}_{\xi\in \mathcal{D}} \hspace{-0.3cm}\log {\prod}_{t=0}^{t_f}\mathbb{P}(\bar{a}_t | s_{t},\bar{\omega}),
\end{align}
\normalsize
where $\bar{a}_t = (a^1_t,\dots,a^n_t)$ is the collection of all agents' actions at time step $t$, $\xi = \{(s_0,\bar{a}_0),\dots,(s_{t_f},\bar{a}_{t_f})\}$ is an  interaction trajectory, and $\mathbb{P}(\bar{a}_t | s_{t}, \bar{\omega})$ denotes the conditional probability of agents' actions given $s_t$ and the current estimate of $\bar{\omega}$. Note that without specifications of the agents' abilities in reasoning about others' reasoning processes, it is hard to further decompose $\mathbb{P}(\bar{a}_t | s_{t}, \bar{\omega})$ to be related to each agent's policy $\pi^i$. Hence, in this work, we explicitly consider agents' bounded intelligence levels that characterize their reasoning abilities and exploit insights from Theory-of-Mind to formulate the multi-agent reward learning problem.

\begin{figure*}[t]
\begin{center}
\begin{picture}(600, 120)
%%%%%%%%%%%%%%%%%%%%%%%%%%%%%%
\put( 0,  0){\epsfig{file=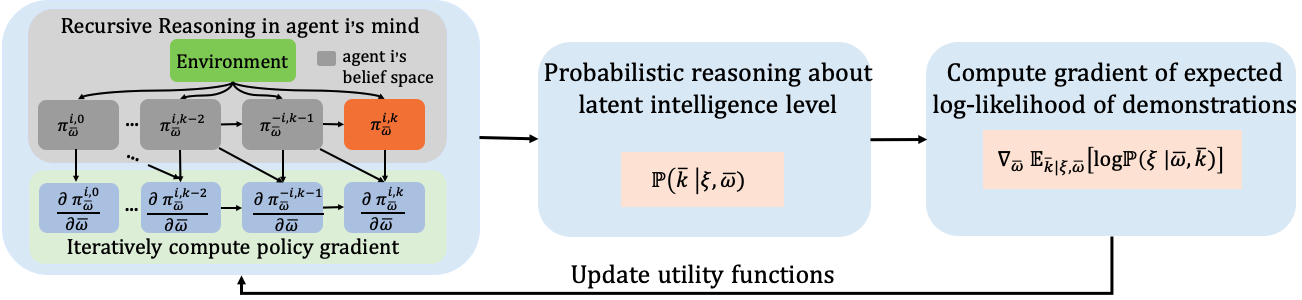,width = 1\linewidth, trim=0.50cm 0.0cm 0.5cm 0cm,clip}}  %%%
\end{picture}
\end{center}
\vspace{-0.5cm}
\caption{Our proposed approach to learning reward functions in multi-agent settings. Left block: procedure for computing ql-$k$ policies and gradients of ql-$k$ policies with respect to weights of reward functions. Middle block: the computed $\pi^{i,k}$ and $\frac{\partial \pi^{i,k}}{\partial \bar{\omega}}$ for each $i$ and $k$ are used to compute the current conditional distribution of $\bar{k}$ given $\xi$ and the current estimate of $\bar{\omega}$. Right block: the information from the previous two blocks are exploited to compute the gradient of the expected likelihood of the demonstrations.}
\label{fig: learning curriculum}
\vspace{-0.3cm}
\end{figure*}

%% file: cognition_modeling.tex
\section{Human cognitive reasoning modelling}\label{sec: ql-k}

Our instance of Theory-of-Mind follows from a recursive reasoning model - quantal level-$k$ model (ql-$k$) \cite{stahl1994experimental} which is shown to be the state-of-the-art \cite{WRIGHT201716} in predicting human strategic behaviors. We briefly introduce the model below.

\noindent
\textbf{Quantal level-$k$ model.} Ql-$k$ model represents humans' recursive reasoning structure in an iterative fashion, starting from ql-$0$ agents who are myopic agents with the lowest intelligence level. Then, a ql-$k$ agent (agent with level-$k$ intelligence), $k\in\mathbb{N}^+$, believes other agents as ql-$(k{-}1)$ agents (i.e., following ql-$(k{-}1)$ policies), and responds accordingly based on such beliefs. Note that with the ql-$0$ model as a base model, the ql-$k$ policies are defined for every agent $i\in\mathcal{P}$ and for every intelligence level $k = 1,\dots,k_{\text{max}}$ through a sequential and iterative process. We illustrate this process from agent $i$'s perspective in the left block of \cref{fig: learning curriculum}.

Specifically, given an initial state $s_t \in \mathcal{S}$, a ql-$k$ agent $i$ maximizes the following objective: $\max_{\pi^{i}} V^{i,k}(s_t)$, where \small $V^{i,k}({s}_t)= \mathbb{E}_{\pi^{-i,k-1}}\Big[ \sum_{\tau=0}^{\infty}\gamma^\tau \mathcal{R}^{i}({s}_{t+\tau}) | \pi^{i}, f \Big]$ \normalsize is the ql-$k$ value function of agent $i$ and \small $\pi^{-i,k-1}$ \normalsize are the predicted ql-$(k-1)$ policies of agents $-i =  \mathcal{P}{\setminus}\{i\}$ from agent $i$'s belief space. It follows that the optimal value function satisfies the following Bellman equation: \small $ V^{*,i,k}({s}) =\max_{a^i\in\mathcal{A}_i} \mathbb{E}_{\pi^{-i,k-1}}\Big[\mathcal{R}^{i}({s}') + \gamma V^{*,i,k}({s}') \big |{s}' = f(s,a^i,a^{-i}), a^{-i} {\sim} \pi^{-i,k-1}\Big]$. \normalsize Then, we define the $Q$-value function as:

\begin{align}\label{equ: quantal Q-value}
    & Q^{*,i,k} (s,a^i) = \mathbb{E}_{\pi^{-i,k-1}} \big[\mathcal{R}^{i}({s}) + \gamma V^{*,i,k}({s}')\Big]\nonumber\\
    & = \mathbb{E}_{\pi^{-i,k-1}} \big[\mathcal{R}^{i}({s}) + \gamma \max_{a^{i,'}} Q^{*,i,k}(s',a^{i,'})\Big].
\end{align}
\normalsize
Note that $Q^{*,i,k}$ is in a form of Bellman equation and can be determined via value iteration. Then we define agent $i$'s ql-$k$ policy using the following quantal best response function:

\begin{align}\label{equ: quantal best response}
\pi^{i,k} ({s}, a^i) =\hspace{-0.1cm} \frac{\text{exp}\big( \lambda^i Q^{*,i,k}({s},a^i)\big)}{\sum_{a'\in \mathcal{A}_{i}} \text{exp}\big(\lambda^i Q^{*,i,k}({s},a')\big)},
\end{align}
where $\lambda^i \in (0,1]$ is the rationality coefficient that controls the degree of agent $i$ conforming to optimal behaviors. Start from the base case, ql-$0$ policies, by sequentially and iteratively solving for $\pi^{i,k} ({s}, a^i)$ via \cref{equ: quantal Q-value} and \cref{equ: quantal best response}, we can compute ql-$k$ policies for an arbitrary $k = 1,2,...$ and for every agent. We note that when agent $i$ predicts its opponents' ql-$(k-1)$ policies, it assumes that its opponents also use rationality coefficient $\lambda^i$ to  establish their ql-$(k-1)$ policies.

\noindent
\textbf{Summary.} Thus far, we have introduced our instance of Theory-of-Mind model exploited to model humans' heterogeneous intelligence levels during strategic interactions and demonstrated how to represent humans' beliefs over other humans' beliefs in an iterative and sequential procedure.

%% file: IRL_method.tex
\section{Cognition-Aware Multi-Agent Inverse Reinforcement Learning}\label{sec: RAMA-IRL}

In this section, we incorporate the ql-$k$ model into our multi-agent IRL formulation.

\noindent
\textbf{Learning with inference on latent intelligence levels.} Though we have developed models that represent humans' cognitive reasoning under different intelligence levels, it is still difficult to further decouple $\mathbb{P}(\bar{a}_t | s_{t}, \bar{\omega})$ in \eqref{equ: maximum likelihood} because humans' intelligence levels are latent cognitive states and can not be directly observed. We assume that humans' true intelligence levels are constant variables in a demonstration $\xi$. We let $\bar{k} = (k^1,\dots,k^n)$ denote the collection of humans' true intelligence levels in $\xi$ and let $\mathbb{K}$ denote a set that contains all possible intelligence levels ($k^i \in \mathbb{K}$). Then, as an approximation, we aim to maximize the expected log-likelihood of the demonstrations conditioned on estimates of the reward parameters $\bar{\omega}$ and a sample $\bar{k}$ with respect to the current conditional distribution of $\bar{k}$ given $\xi$ and the current estimates of $\bar{\omega}$:
\begin{align}\label{equ: maximum likelihood with latent state}
    &\max_{\bar{\omega}} {\sum}_{\xi\in \mathcal{D}} \hspace{-0.3cm}\log \mathbb{P}\left(\xi | \bar{\omega}\right)\nonumber\\
    &\approx
	 \max_{\bar{\omega}} {\sum}_{\xi\in \mathcal{D}} \sum_{\bar{k} \in \mathbb{K}^n} \log\big( \mathbb{P}(\xi | \bar{\omega},\bar{k})\big) \mathbb{P}(\bar{k}|\xi,\bar{\omega}),\nonumber\\
	& \mathbb{P}(\xi | \bar{\omega},\bar{k}) =  \prod_{t=0}^{t_f}\prod_{i=1}^{n} \pi^{i,k^i}_{\bar{\omega}}\hspace{-0.1cm}(s_t,a^i_t), \mathbb{P}(\bar{k}|\xi,\bar{\omega}) = \prod_{i=1}^{n} \mathbb{P}(k^i|\xi,\bar{\omega}),
	%& \log \big(\mathbb{E}_{\bar{k}|\xi,\bar{\omega}}\mathbb{P}(\xi | \bar{\omega},\bar{k})\big) \nonumber\\
	%& = {\sum}_{t=0}^{N-1}\hspace{-0.2cm} \log{\sum}_{\bar{k} \in \mathbb{K}^n} \prod_{i=1}^n \pi^{i,k^i}_{\bar{\omega}}\hspace{-0.1cm}(s_t,a^i_t)\mathbb{P}(k^i | \xi, \bar{\omega} ),
\end{align}
% The expected value of the likelihood of one demonstration $\mathbb{E}_{\bar{k}|\xi,\bar{\omega}} \big[ \log \mathbb{P}(\xi | \bar{\omega},\bar{k})\big]$ is computed as follows:
% \begin{align}\label{equ: trajectory likelihood full}
%     & \mathbb{E}_{\bar{k}|\xi,\bar{\omega}} \big[ \log \mathbb{P}(\xi | \bar{\omega},\bar{k})\big] = \sum_{\bar{k} \in \mathbb{K}^n} \log\big( \mathbb{P}(\xi | \bar{\omega},\bar{k})\big) \mathbb{P}(\bar{k}|\xi,\bar{\omega}), \nonumber\\
%     & \mathbb{P}(\xi | \bar{\omega},\bar{k}) =  \prod_{t=0}^{t_f}\prod_{i=1}^{n} \pi^{i,k^i}_{\bar{\omega}}\hspace{-0.1cm}(s_t,a^i_t),\nonumber\\
%     & \mathbb{P}(\bar{k}|\xi,\bar{\omega}) = \prod_{i=1}^{n} \mathbb{P}(k^i|\xi,\bar{\omega}),
% \end{align}
where $\pi^{i,k^i}_{\bar{\omega}}$ denotes the ql-$k^i$ policy of agent $i$ and $\mathbb{P}(k^i | \xi, \bar{\omega} )$ denotes the estimated probability distribution of agent $i$'s intelligence level conditioned on the current demonstration and estimate of reward parameters, which can be computed by applying recursive Bayesian inference along $\xi$:

\vspace{-0.4cm}
\small
\begin{align}\label{equ: level inference}
    \mathbb{P}(k^i = k | \xi_t, \bar{\omega})= \frac{\pi^{i,k}_{\bar{\omega}}(s_t,a_t^i) \mathbb{P}(k^i = k | \xi_{t-1}, \bar{\omega})}{\sum_{k' \in \mathbb{K}} \pi^{i,k'}_{\bar{\omega}}(s_t,a_t^{i},) \mathbb{P}(k^i = k' | \xi_{t-1}, \bar{\omega})},
\end{align}
\normalsize
where $\xi_t = \{(s_0,\bar{a}_0),\dots,(s_{t},\bar{a}_{t})\}$ ($t\leq t_f$) and the initial prior distribution is a uniform distribution over $\mathbb{K}$. Then, we set $ \mathbb{P}(k^i = k | \xi, \bar{\omega}) =  \mathbb{P}(k^i = k | \xi_{t_f}, \bar{\omega})$.

In summary, in order to evaluate the likelihood of a demonstration induced by the current estimate of agents' reward parameters using the ql-$k$ model with the latent variable $\bar{k}$, we first compute the current conditional distribution of $\bar{k}$ given $\xi$ and the current estimate of agents' reward parameters $\bar{\omega}$ ($\mathbb{P}(k^i | \xi, \bar{\omega} )$), then compute the expected likelihood of the demonstrations with respect to $\mathbb{P}(k^i | \xi, \bar{\omega} )$. After that, the gradient of the objective function \cref{equ: maximum likelihood with latent state} can be used to find a locally optimal $\bar{\omega}$. Such a learning procedure is illustrated in \cref{fig: learning curriculum}.

\noindent
\textbf{Analytic Q-value gradient approximation.} It follows from \cref{equ: maximum likelihood with latent state} that the gradient of the learning objective function with respect to $\bar{\omega}$ depends on the gradients of $\pi^{i,k}_{\bar{\omega}}$ and $\mathbb{P}(k | \xi, \bar{\omega} )$ with respect to $\bar{\omega}$, which both further depend on the gradient of $Q^{i,k}_{\bar{\omega}}$ with respect to $\bar{\omega}$ according to \cref{equ: quantal best response}. Due to the non-differentiable max operator associated with the computation of $Q^{i,k}_{\bar{\omega}}$ in \cref{equ: quantal Q-value}, we utilize a smooth approximation of the $Q$ value:
\vspace{-0.2cm}
\begin{align}
    Q^{i,k}_{\bar{\omega}}(s,a^i)   \approx &\sum_{a^{-i}} \mathbb{P}(a^{-i}|s,\bar{\omega}) \Big( \mathcal{R}^i_{\bar{\omega}}(s') \nonumber \\
    & + \gamma\big(\sum_{a^{i,'}}(Q^{i,k}_{\bar{\omega}}(s',a^{i,'}))^{\kappa}\big)^{\frac{1}{\kappa}}\Big),
\end{align}
where the parameter $\kappa$ controls the approximation error, and when $\kappa \rightarrow \infty$, the approximation becomes exact.

Let us assume that we have access to $\frac{\pi^{i,k-1}_{\bar{\omega}}}{\partial \bar{\omega}}$ of each agent, then the gradient of the value function $Q^{i,k}_{\bar{\omega}}$ with respect to $\bar{\omega}$ can be approximated as follows:

\vspace{-0.2cm}
\small
\begin{align}\label{equ: Q-gradient-bellman}
    & \frac{Q^{i,k}_{\bar{\omega}}}{\partial \bar{\omega}}(s,a^i)  \approx \sum_{a^{-i}} \Bigg[\frac{\partial \Pi^{-i,k-1}}{\partial \bar{\omega}}(a^{-i}|s,\bar{\omega}) \Bigg( \mathcal{R}^i_{\bar{\omega}}(s') \nonumber \\
    & \hspace{+1cm} + \gamma\Big(\sum_{a^{i,'}}\big(Q^{i,k}_{\bar{\omega}}(s',a^{i,'})\big)^{\kappa}\Big)^{\frac{1}{\kappa}}\Bigg) \nonumber \\
    & 
    + \mathbb{P}(a^{-i}|s,\bar{\omega})\Big( \frac{\partial \mathcal{R}^{i}_{\bar{\omega}}}{\partial \bar{\omega}}(s') + \gamma\frac{1}{\kappa}\big(\sum_{a^{i,'}}(Q^{i,k}_{\bar{\omega}}(s',a^{i,'}))^{\kappa}\big)^{\frac{1-\kappa}{\kappa}}\nonumber\\
   & \quad \cdot \sum_{a^i}  \kappa \Big( Q^{i,k}_{\bar{\omega}}(s',a^i)\Big)^{\kappa{-}1} \frac{Q^{i,k}_{\bar{\omega}}}{\partial \bar{\omega}}(s',a^i) \Bigg],
\end{align}
\normalsize

where $\frac{\partial \Pi^{-i,k-1}}{\partial \bar{\omega}}(a^{-i}|s,\bar{\omega})$ can be computed as follows:
\begin{align}
&\frac{\partial \Pi^{-i,k-1}}{\partial \bar{\omega}}(a^{-i}|s,\bar{\omega}) \nonumber\\
& = \sum_{j\in \mathcal{P}\setminus \{i\}} \hspace{-0.2cm} \frac{\pi^{j,k-1}}{\partial \bar{\omega}}(s,a^j|\bar{\omega})\prod_{e\in \mathcal{P}\setminus\{i,j\}} \hspace{-0.2cm}\pi^{e,k-1}_{\bar{\omega}}(s,a^e|\bar{\omega}),
\end{align}
where $\frac{\pi^{j,k-1}_{\bar{\omega}}}{\partial \bar{\omega}}$ is assumed to be known. Note that \cref{equ: Q-gradient-bellman} expresses the gradient of agent $i$'s ql-$k$ $Q$-value function with respect to $\bar{\omega}$ in a form a Bellman equation that only depends on agent $i$'s ql-$k$ Q-value function, the gradients of agents' reward functions with respect to $\bar{\omega}$, agents $-i$'s ql-$(k-1)$ policies and the corresponding policy gradients, which we all have access to. Therefore, we can compute $ \frac{\partial Q^{i,k}_{\bar{\omega}}}{\partial \bar{\omega}}$ via value iteration algorithm and then $\frac{\partial \pi^{i,k}}{\partial \bar{\omega}}$ can also be computed accordingly by differentiating \cref{equ: quantal best response}.

\input{learning_algo}

\noindent
\textbf{Policy gradient.} In the previous part, we show that with each agent's $\frac{\partial \pi^{i,k-1}}{\partial \bar{\omega}}$ known, then we can compute  $\frac{\partial \pi^{j,k}}{\partial \bar{\omega}}$ for each agent $j \in \mathcal{P}$. Therefore, starting from $\frac{\partial \pi^{i,0}}{\partial \bar{\omega}}$ of each agent, we can compute $\frac{\partial \pi^{i,k}}{\partial \bar{\omega}}$ through a sequential and iterative process similar to the procedure for computing $\pi^{i,k}$ desired be \cref{sec: ql-k}. Since ql-$0$ agents (agents with the lowest intelligence level) are normally represented as reflexive agents who do not consider sophisticated interactions with others \cite{wright2014level}, $\frac{\partial \pi^{i,0}}{\partial \bar{\omega}}$ can be computed straightforwardly based on a particular instantiation. The procedure for computing $\frac{\partial \pi^{i,k}}{\partial \bar{\omega}}$ is illustrated in lower-left block in \cref{fig: learning curriculum}.

With $\frac{\partial \pi^{i,k}}{\partial \bar{\omega}}$ computed, the gradient $\frac{\partial \mathbb{P}(k^i|\xi,\bar{\omega})}{\partial \bar{\omega}}$ can be obtained by differentiating \cref{equ: level inference}, which yields a recursive format from time 0 to time $t$ and can be easily computed with initialization $\frac{\partial \mathbb{P}(k^i|\xi_0,\bar{\omega})}{\partial \bar{\omega}} = \textbf{0}$.

\noindent
\textbf{Learning algorithm.} With both $\frac{\partial \pi^{i,k}}{\partial \bar{\omega}}$ and $\frac{\partial \mathbb{P}(k^i|\xi,\bar{\omega})}{\partial \bar{\omega}}$ available, the gradient of the learning objective function \cref{equ: maximum likelihood with latent state} with respect to $\bar{\omega}$ can be computed straightforwardly and used to update the estimate of $\bar{\omega}$. In practice, we also regularize the reward parameters during learning. The overview of learning curriculum is illustrated in \cref{fig: learning curriculum} and our Cognition-Aware Multi-Agent Inverse Reinforcement Learning algorithm is summarized in  \cref{alg: RA-MAIRL}.

%% file: learning_algo.tex
%\vspace{-0.3cm}
\begin{algorithm}
	\caption{Cognition-Aware Multi-Agent Inverse Reinforcement Learning algorithm}
	\label{alg: RA-MAIRL}
	\textbf{Input:} A demonstration set $\mathcal{D}$ and learning rate $\eta$
	
	\textbf{Output:} Learned parameters $\bar{\omega}$.
	
	Initialize $\bar{\omega}$.
	
	\While{\text{not converged}}{
		
		Compute $\pi^{i,k}_{\bar{\omega}}$ and $\frac{\partial \pi^{i,k}}{\partial \bar{\omega}}$ for each $i$ and $k$;
		
		Initialize $\nabla_{\bar{\omega}}$;
		
		\For{ $\xi \in \mathcal{D}$}{
		    Estimate agents' latent intelligence levels in $\xi$ via \cref{equ: level inference};
		    
		Compute gradient of the expected log-likelihood of the demonstrations following: $\nabla_{\bar{\omega}} + = \frac{ \partial \Big(\sum_{\bar{k} \in \mathbb{K}^n} \log\big( \mathbb{P}(\xi | \bar{\omega},\bar{k})\big) \mathbb{P}(\bar{k}|\xi,\bar{\omega}) \Big)}{\partial \bar{\omega}}$;
		}
		
		Update the parameters following: $\bar{\omega} = \bar{\omega} + \eta \nabla_{\bar{\omega}}$;
	}
	
	\textbf{Return:} $\bar{\omega}$
\end{algorithm}

%% file: result.tex
\section{Experiment Design}\label{sec: experiment design}

In this section, we design two experiments to validate our hypothesis: \textit{reasoning about humans’ latent intelligence levels adds flexibility to inverse learning algorithm and helps learn reward functions that explain and reconstruct human driving behaviors better}.

\subsection{Environments}
Although our formulation and approach generally apply to settings with multiple agents ($n\geq2$), here we focus on the cases where the ego agent interacts with another opponent agent in order to compare with baselines that are tailored for pair-wise relationships.

\begin{figure}[ht]
\begin{center}
\begin{picture}(300, 120)
%%%%%%%%%%%%%%%%%%%%%%%%%%%%%%
\put( 20,  50){\epsfig{file=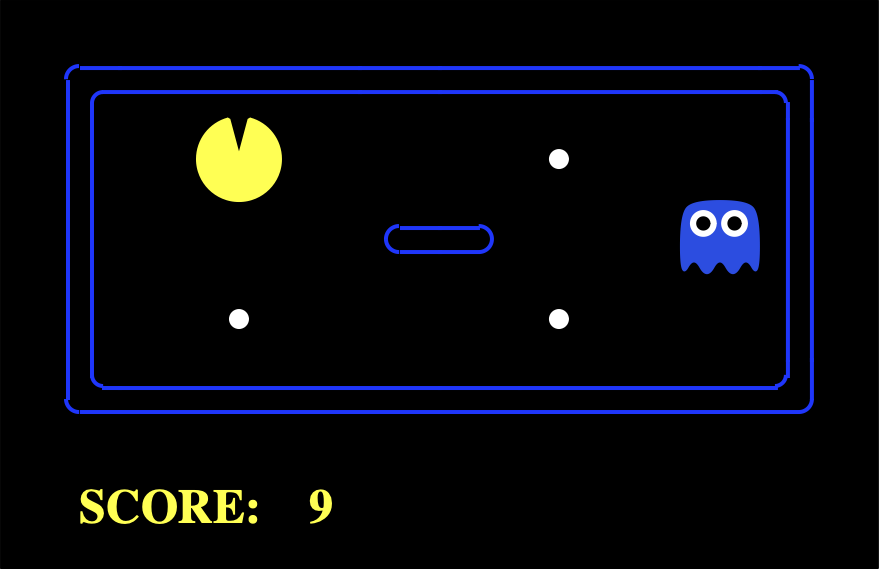,width = 0.8\linewidth, height = 0.3\linewidth, trim=2.0cm 6.0cm 2cm 2cm,clip}}  %%%

\put( 20,  0){\epsfig{file=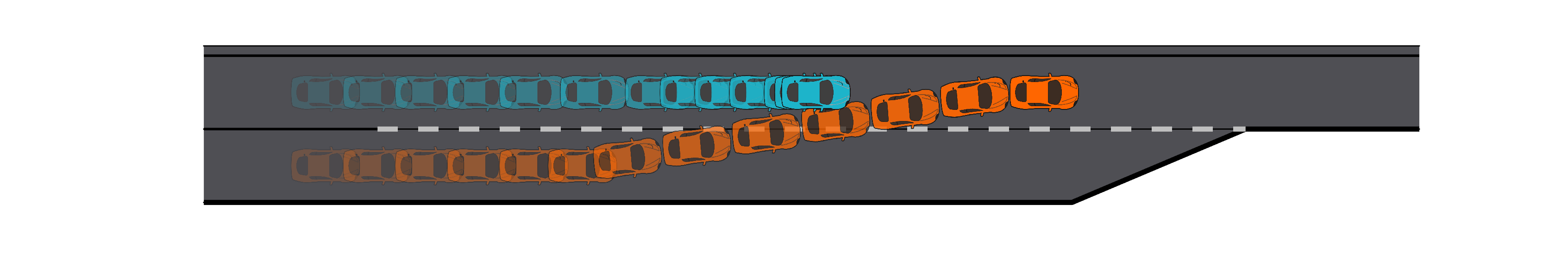,width = 0.8\linewidth, trim=5.0cm 0.0cm 4cm 0.5cm,clip}}  %%%
\put(0,110){(a)}
\put(0,40){(b)}
\end{picture}
\end{center}
\vspace{-0.8cm}
\caption{Simulated environments. (a): Pac-man aims to collect all the dots and avoid being eaten by the ghost. (b): The lower-lane car wants to effectively execute the merging while remaining safe.}
\label{fig: environments}
\vspace{-0.3cm}
\end{figure}

\noindent
\textbf{Zero-sum game.} We consider a $2D$ maze game, Pac-Man (as shown in \cref{fig: environments} (a)), in which the pac-man aims to collect all the pac-dots in the shortest time possible and the ghost aims to prevent the pac-man from doing so. The agents share the same action set that contains five actions: moving up, down, left, right, or stay. The state of the game is defined as $s = [x^1, y^1, x^2, y^2, \bar{\mathbb{I}}]$, where $x$ ($y$) is the longitudinal (lateral) position, the superscript $1$ ($2$) denotes the pac-man (ghost), and $\bar{\mathbb{I}} = [\mathbb{I}^1,\dots,\mathbb{I}^{n_d}]$ is a vector of Booleans that track the validity of the $n_d$ pac-dots in the game.

\noindent
\textbf{General-sum game.} We use a driving environment that consists of two cars on a two-lane road. As shown in \cref{fig: environments} (b), the orange car has to merge to its adjacent (upper) lane which is occupied by the blue car. Sophisticated and interactive behaviors are likely to happen under such a scenario \cite{choudhury2019utility}. The dynamics of the driving scenario are represented as
$\left[ \begin{matrix}\dot{x}^{1} & \dot{y}^{1} & \dot{x}^{2} & \dot{v}^{1} & \dot{v}^{2}\end{matrix}\right] = \left[\begin{matrix}v^{1} & w^{1} & v^{2} & a^{1} & a^{2}\end{matrix} \right]$, where $x$ ($y$) is the longitudinal (lateral) position, $v$ ($w$) is the longitudinal (lateral) speed, $a$ is the longitudinal acceleration, and the superscript $1$ ($2$) denotes the lower-lane (upper-lane) car. The sampling period is $\Delta t = 0.5[s]$. We use a state grid of size $50\times4\times50\times5\times5$ to represent the discrete states of the game in a similar way as in \cite{fisac2019}.

\subsection{Manipulated Variables}
The manipulated variable is the choice of different inverse reinforcement learning algorithms for learning agents' reward functions from demonstrations. In addition to our proposed approach, we consider two baseline algorithms that are both extensions of MaxEnt IRL in two-agent settings. In contrast to our approach, they both assign agents' roles/strategies \textit{a priori} during learning.

\noindent
\textbf{PR2-MaxEnt IRL.} Probabilistic recursive reasoning (PR2) framework \cite{wen2019probabilistic} is a particular instantiation of the quantal level-$k$ model. However, PR2 assumes all agents perform depth-$1$ recursive reasoning. As the first baseline, we use the PR2 framework to model interactions and plug the PR2-$Q$ algorithm into our learning framework: {we assume that all humans in a demonstration perform depth-$1$ recursive reasoning when interacting with others and we use the PR2-$Q$ algorithm to generate humans' policies conditioned on the roll-out reward functions. Then, we use the generated policies to compute the gradient of the log-likelihood of demonstrations ( \cref{equ: maximum likelihood}) without reasoning about humans' latent cognitive states since the PR2 framework assigns humans' intelligence levels \textit{a priori}.}

\noindent
\textbf{MaxEnt IRL with leader-follower (LF) roles.} Our second baseline (LF-MaxEnt IRL) is the MaxEnt IRL algorithm used in \cite{sadigh2016planning, sadigh2017active,sun_probabilistic_2018,suncourteous}, which pre-assigns the leader and the follower during learning (refer to \cref{sec: related works} for more details).

%\vspace{-0.2cm}
\subsection{Dependent Measures}
{In order to validate our proposed approach and compare it against the baselines, we conducted two experiments. In the first experiment, we utilize demonstrations from synthetic agents in the two environments, then we can quantify the performance of IRL algorithms in terms of: (a) the correlations between the ground-truth reward functions (handcrafted) and the learned reward functions; (b) the log-likelihood of the demonstrations induced by an IRL algorithm. In the second experiment, we further evaluate the effectiveness different IRL algorithms in the driving domain with real traffic data by measuring: (a) the log-likelihood of the demonstrations induced by an IRL algorithm; (b) the similarity between the ground-truth interaction trajectories in test set and the trajectories generated using the reward functions learned by an IRL algorithm and its interaction model.}

%\vspace{-0.2cm}
\subsection{Feature Selection}
To avoid confounding variables, we control the features associated with agents' reward functions in both environments. More specifically, we let agents optimize over a linear combination of common features in our approach and the baselines. The feature weights are our reward parameters $\bar{\omega}$.

\noindent \textbf{Pac-Man environment.} Pac-man considers the following features: 1) \textit{Safety}: this feature represents the pac-man's preference to avoid the ghost and is defined as $f_{1,p} {=} \exp(d)$, where $d$ denotes the $L{-}1$ distance between the pac-man and the ghost. 2) \textit{Proximity}: this feature represents how the pac-man wants to reach the closest pac-dot and is defined as $f_{2,p} = \frac{1}{l_{d,min}}$, where $l_{d,min}$ is the distance from the pac-man to the closet pac-dot. 3) \textit{Winning}: this feature represents how the pac-man wants to eat more pac-dots for winning the game and is defined as $f_{3,p} = \frac{1}{n_{ud}}$, where $n_{ud}$ denotes the number of uncollected pac-dots. {The ghost considers similar features as the pac-man does but with modifications}: $f_{1,g} {=} \frac{1}{d}$, $f_{2,g} {=} f_{2,p}$, and $f_{3,g} {=} \frac{1}{f_{3,p}}$.

\noindent \textbf{Driving environment.} For the driving domain, we select features based on previous works \cite{choudhury2019utility, suncourteous}: 1) \textit{Progress}: this feature represents humans' preference to drive faster and merge to the desired lane. It is defined as $f_1 {=} (\frac{v}{v_{max}})^2 {+} (\frac{y_{eff}}{|y_l {-} y_0|})^2$, where $v_{max}$ denotes the maximum allowable speed, $y_0$ denotes the initial lateral position of the car, $y_l$ denotes the lateral coordinate of the target lane, and $y_{eff}$ denotes the effective transnational distance that the car has executed towards the target lane ($max(\frac{y_{eff}}{|y_l {-} y_0|}) {=} 1$ and $min(\frac{y_{eff}}{|y_l - y_0|}) {=} 0$). 2) \textit{Comfort}: this feature represents humans' desire to operate smoothly and is defined as $f_2 {=} (\frac{a}{a_{max}})^2$. 3) \textit{Safety}: this feature represents how humans want to avoid collisions, and is defined as $f_3 = \sum_{i=1}^{n_o} (\frac{min(d_i,d_{safe})}{d_{safe}})^2$, where $d_{safe}$ is a pre-defined safety distance and $d_i$ is the distance between the ego agent and the $i$-th obstacle.

\section{Results and Analysis}
\label{sec:experiment_results}
\subsection{Performance with Demonstrations from Synthetic Agents}
\noindent
\textbf{Synthetic agents.} In both environments, we model the synthetic agents as ql-$k$ agents with various $k\in\mathbb{K}$. Recall that policies of ql-$0$ agents are required to initiate the recursive reasoning process \cref{equ: quantal Q-value}. In line with previous works \cite{li2017game, bouton2020reinforcement, tian2020game}, we let ql-$0$ agents be reflective (non-strategic) agents who do not explicitly take into account their opponents' possible responses but rather maximize their immediate rewards by treating other agents as stationary objects. The ground-truth reward functions of synthetic agents in each environment are manually tuned to achieve reasonable behaviors. For each environment, we generate $30$ interactions with random initial states and ground-truth agent types ($k$).

\vspace{-0.0cm}
\begin{figure}[ht]
\begin{center}
\begin{picture}(300, 110)
%%%%%%%%%%%%%%%%%%%%%%%%%%%%%%
\put(0,  0){\epsfig{file=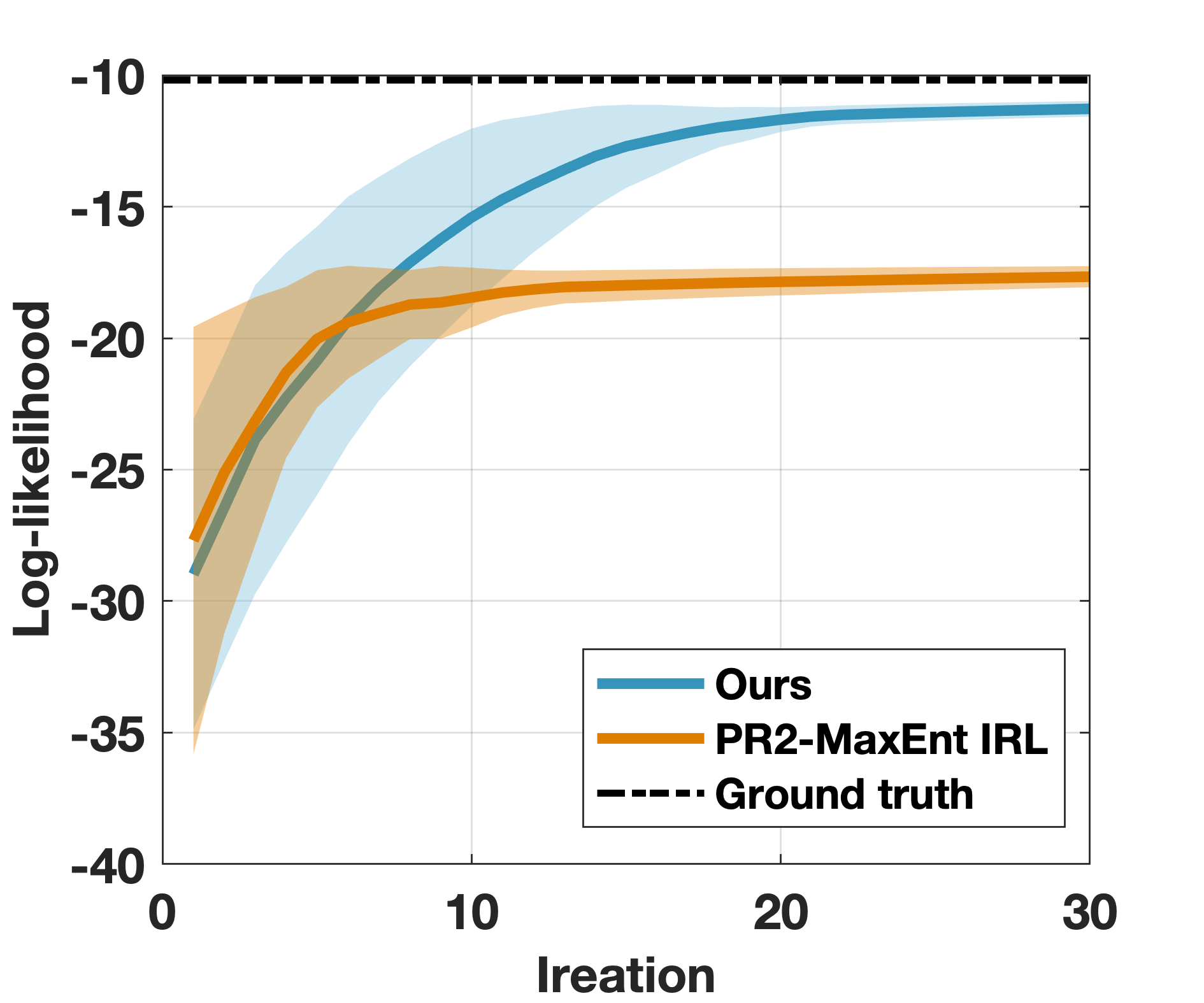, width = 0.5\linewidth, trim=0.0cm 0.0cm 0.4cm 0.5cm,clip}}  %%%
\put(120,  0){\epsfig{file=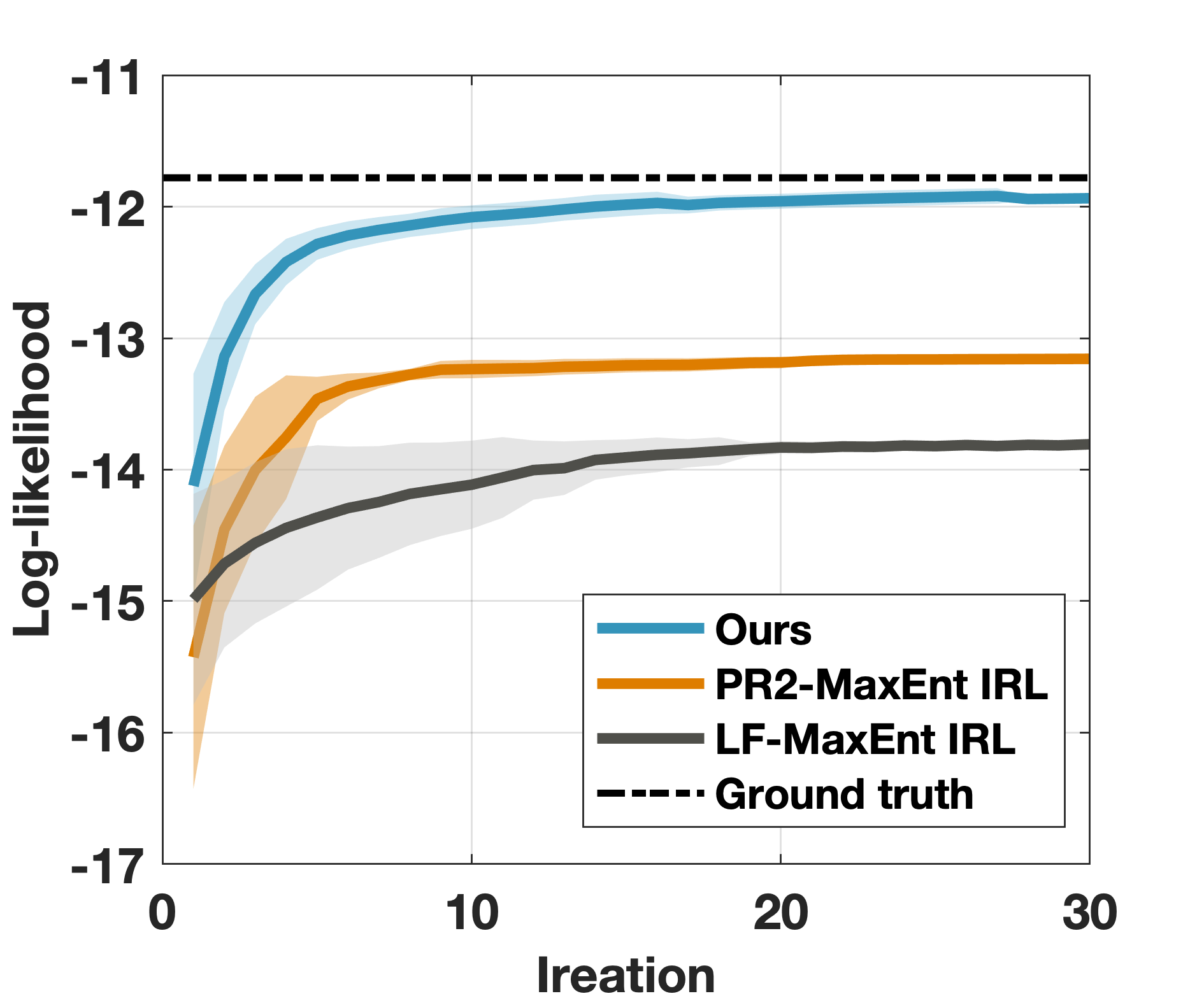, width = 0.5\linewidth, trim=0.0cm 0.0cm 0.4cm 0.5cm,clip}}  %%%
% \put(0,130){(a)}
% \put(0,40){(b)}
\end{picture}
\end{center}
\vspace{-0.3cm}
\caption{Histories of the log-likelihood of the demonstration set during learning (solid line represents the mean and the light shaded area represents the 95\% confidence tube of the data). Left: Pac-Man. Right: Driving.}
\label{fig: simulated_human_loss_hist}
\vspace{-0.2cm}
\end{figure}

\begin{figure}[ht]
\begin{center}
\begin{picture}(300, 70)
%%%%%%%%%%%%%%%%%%%%%%%%%%%%%%
\put(0,  0){\epsfig{file=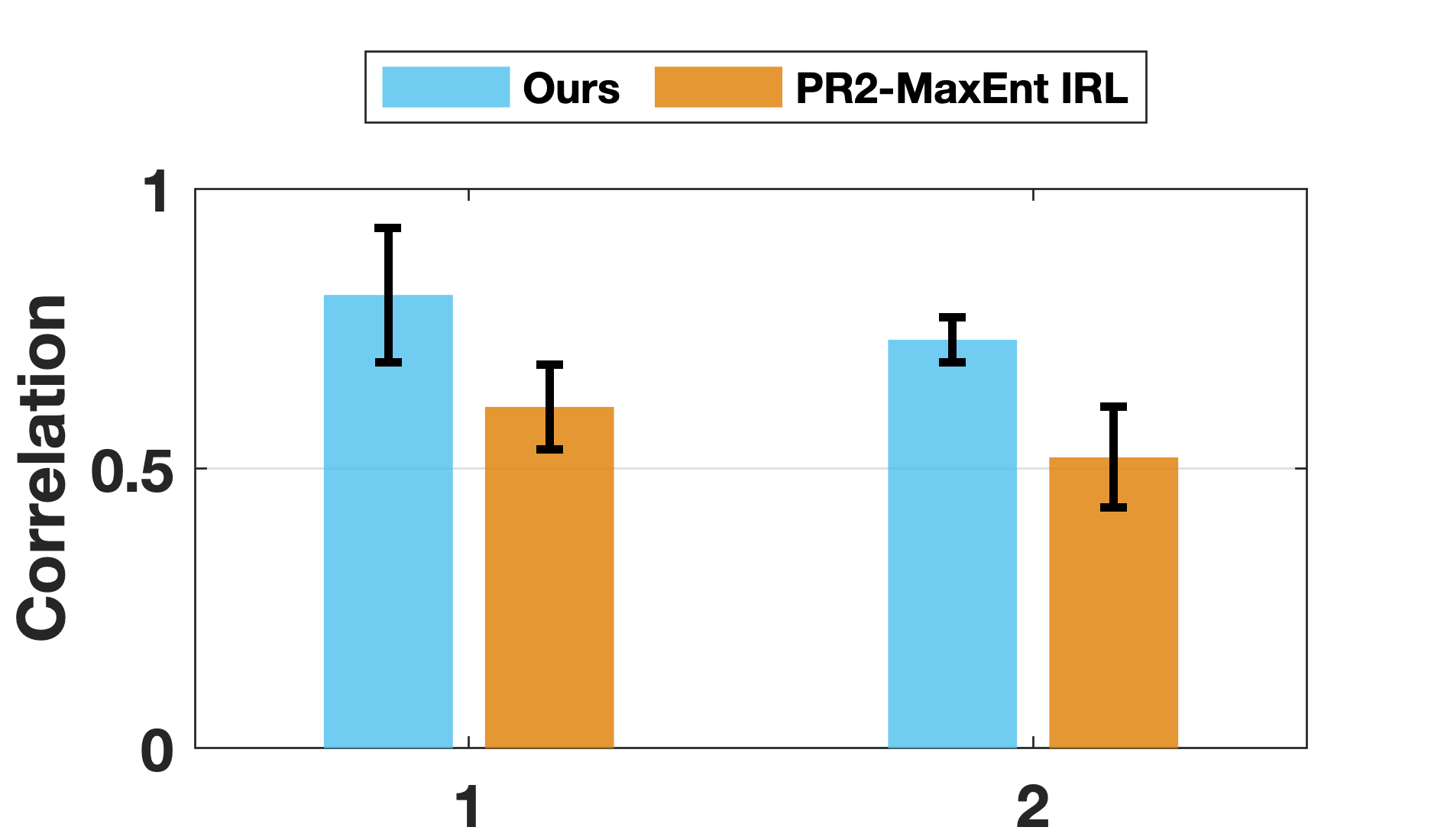, width = 0.5\linewidth, trim=0.0cm 0.50cm 0.4cm 0.5cm,clip}}  %%%
\put(120,  0){\epsfig{file=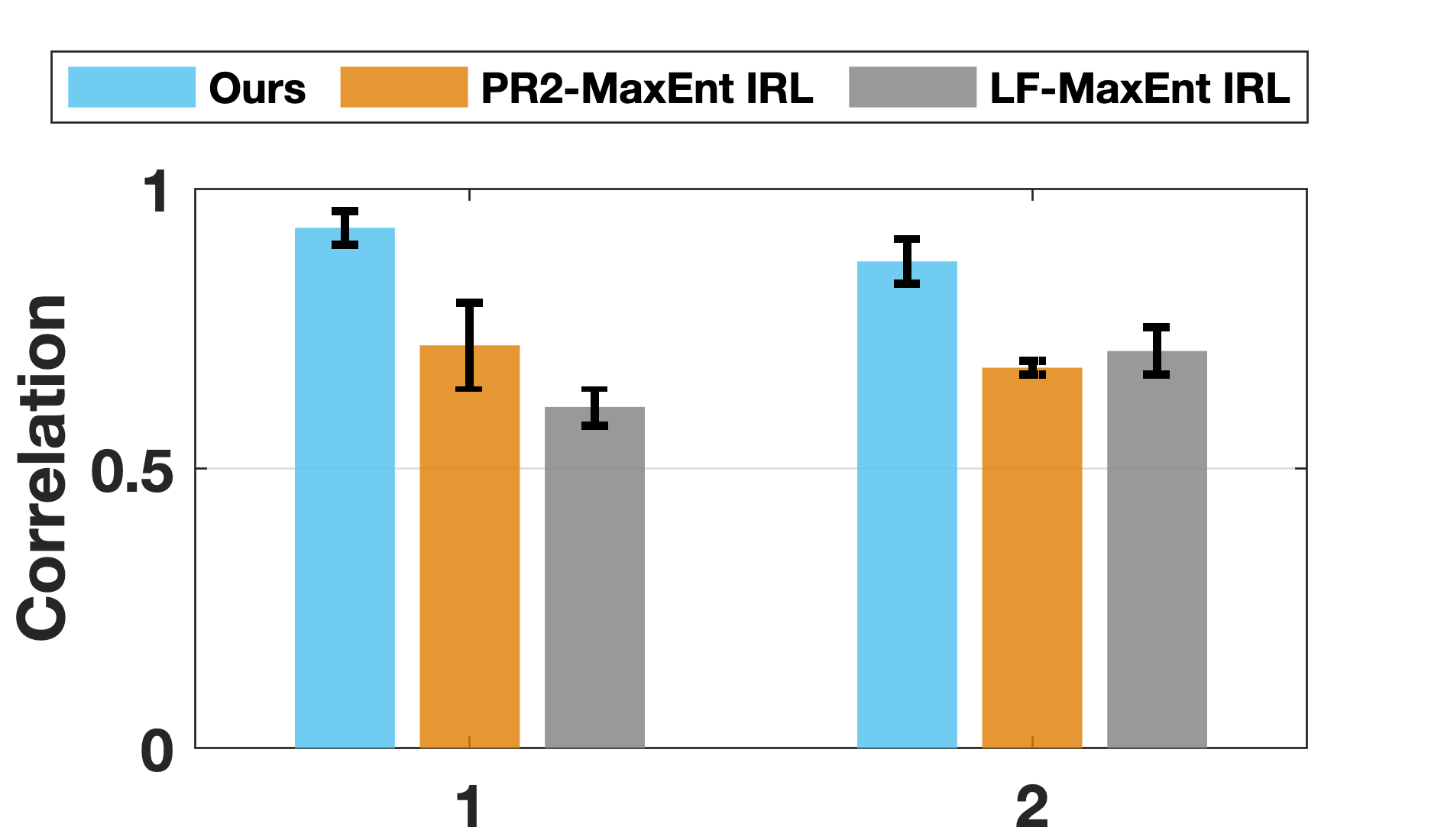, width = 0.5\linewidth, trim=0.0cm 0.5cm 0.4cm 0.5cm,clip}}  %%%
\small
\put(35,-5){SCC}
\put(85,-5){PCC}
\put(155,-5){SCC}
\put(205,-5){PCC}
\normalsize
\end{picture}
\end{center}
\vspace{-0.3cm}
\caption{Statistical correlations between the learned reward parameters and the ground-truth ones. Left: Pac-Man. Right: Driving.}
\label{fig: simulated_human_corr}
\vspace{-0.2cm}
\end{figure}
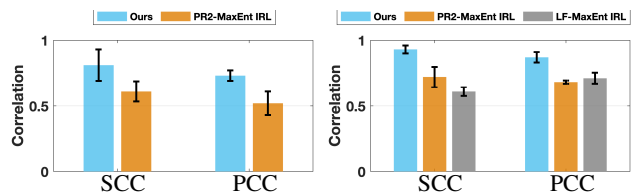

\noindent
\textbf{Learning performance.} In \cref{fig: simulated_human_loss_hist}, we show the histories of the log-likelihood of the demonstration set (learning objective function) using our algorithm in each environment (blue line). The black lines denote the ground truth log-likelihood of the demonstration set evaluated using the models of synthetic agents. It can be observed that the log-likelihood of the demonstration set induced by the learned reward functions approaches to the ground-truth value as the learning algorithm converges. In \cref{fig: simulated_human_corr}, we compare the learned reward parameters and the ground-truth reward parameters using Pearson’s correlation coefficient (PCC) and Spearman’s rank correlation coefficient (SCC). In general, a higher PCC indicates a higher linear correlation, and a higher SCC represents a stronger monotonic relationship. It can be observed that our approach is able to recover reward parameters with a high linear correlation and a strong monotonic relationship to the ground-truth ones.

\vspace{-0.2cm}

\noindent
\textbf{Remark.} In \cref{fig: simulated_human_loss_hist} and \cref{fig: simulated_human_corr}, we also plot the results with the baselines. Such a comparison with the baselines mainly serves as a sanity check: with synthetic agents, our approach is supposed to work better than the baselines since the interaction models of synthetic agents align with our approach to modeling interactions during learning. However, this experiment illustrates the effectiveness of our algorithm in both adversarial and cooperative settings if agents were indeed following our Theory-of-Mind model for making decisions. In the following section, we utilize real human driving data to test the performance of our approach and make a detailed comparison against the baselines.

\subsection{Learning Driver Reward Functions from Traffic Data}
\noindent
\textbf{Traffic data.} We extract real driving data in a forced merging scenario (DRCHNMergingZS) from INTERACTION dataset\cite{interactiondataset}. We extract 30 interactions (sampling time: $0.5 [s]$) as the training set and another 15 interactions as the test set.

\begin{figure}[ht]
\begin{center}
\begin{picture}(300, 90)
%%%%%%%%%%%%%%%%%%%%%%%%%%%%%%
\put(-5,  -3){\epsfig{file=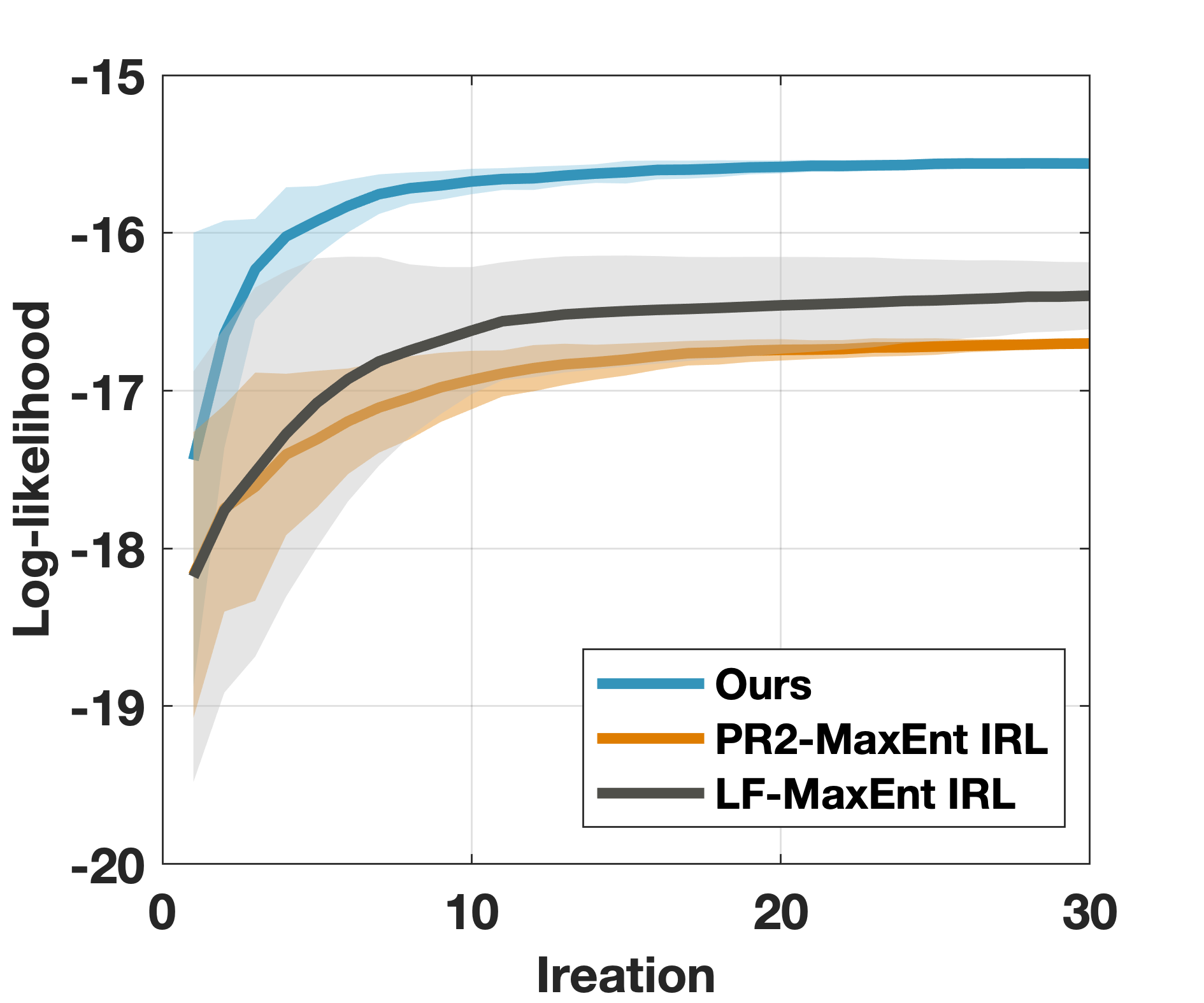, width = 0.45\linewidth, trim=0.0cm 0.0cm 0.4cm
0.5cm,clip}}  %%%
\put(105,  61){\epsfig{file=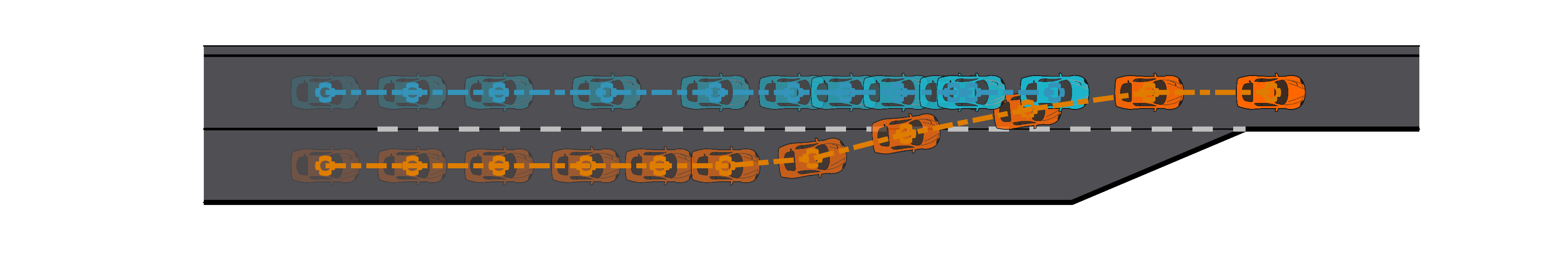, width = 0.58\linewidth, trim=4.0cm 0.5cm 4cm 0.7cm,clip}}
\put(105,  39){\epsfig{file=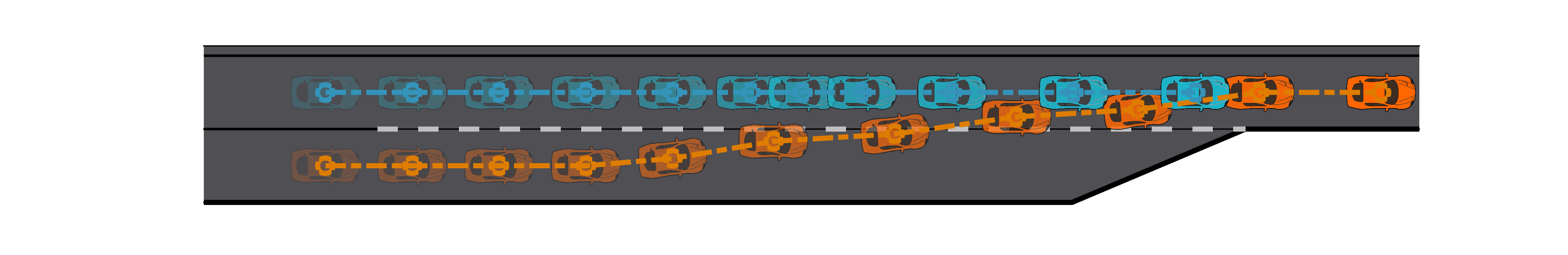, width = 0.58\linewidth, trim=4.0cm 0.5cm 4cm 0.7cm,clip}}
\put(105,  17){\epsfig{file=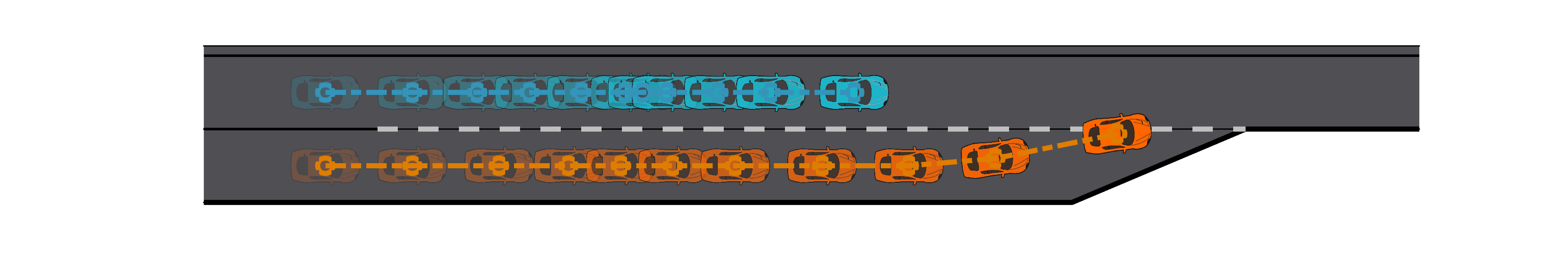, width = 0.58\linewidth, trim=4.0cm 0.5cm 4cm 0.7cm,clip}}
\put(105,  -5){\epsfig{file=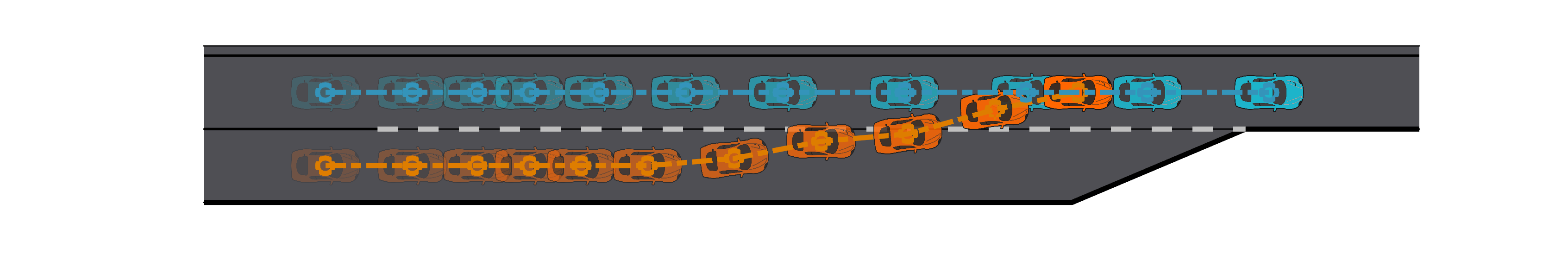, width = 0.58\linewidth, trim=4.0cm 0.5cm 4cm 0.7cm,clip}}
\put(235,64){(a)}
\put(235,42){(b)}
\put(235,20){(c)}
\put(235,-2){(d)}
\end{picture}
\end{center}
\vspace{-0.5cm}
\caption{ Left: Histories of the log-likelihood of the demonstration set during learning. Right: An example of test interaction trajectory and reconstructed interaction trajectories (a: ground truth; b: Ours; c: PR2-MaxEnt IRL; d: LF-MaxEnt IRL).}
\label{fig: real_human_loss_hist}
\end{figure}
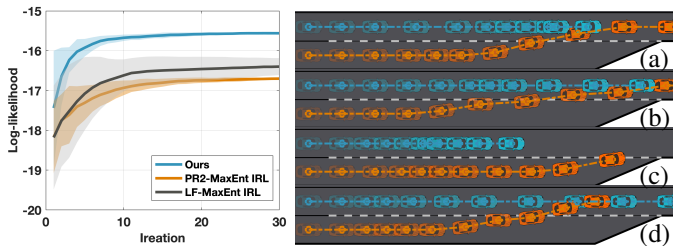

\noindent
\textbf{Learning performance.} In the left plot of \cref{fig: real_human_loss_hist}, we show the histories of the log-likelihood of the demonstrations during learning. It can be observed that our approach can better explain human driving behaviors, yielding a higher likelihood for the demonstration set based on the learned reward functions and the exploited interaction model. More importantly, our approach performs better at very early iterations, which indicates that, by reasoning about human drivers' latent cognitive state (intelligence level), the learning algorithm is offered more flexibility to explain human behaviors. 

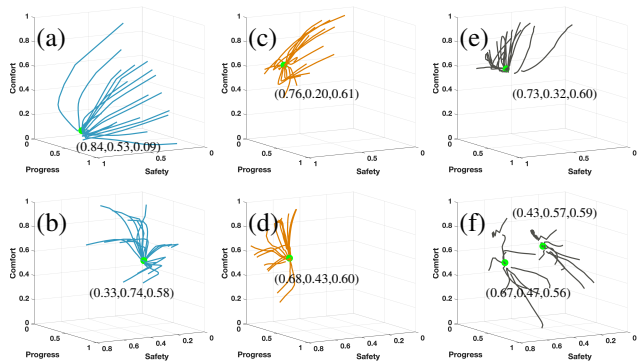
\begin{figure}[ht]
\begin{center}
\begin{picture}(300, 140)
%%%%%%%%%%%%%%%%%%%%%%%%%%%%%%
\put(0,  70){\epsfig{file=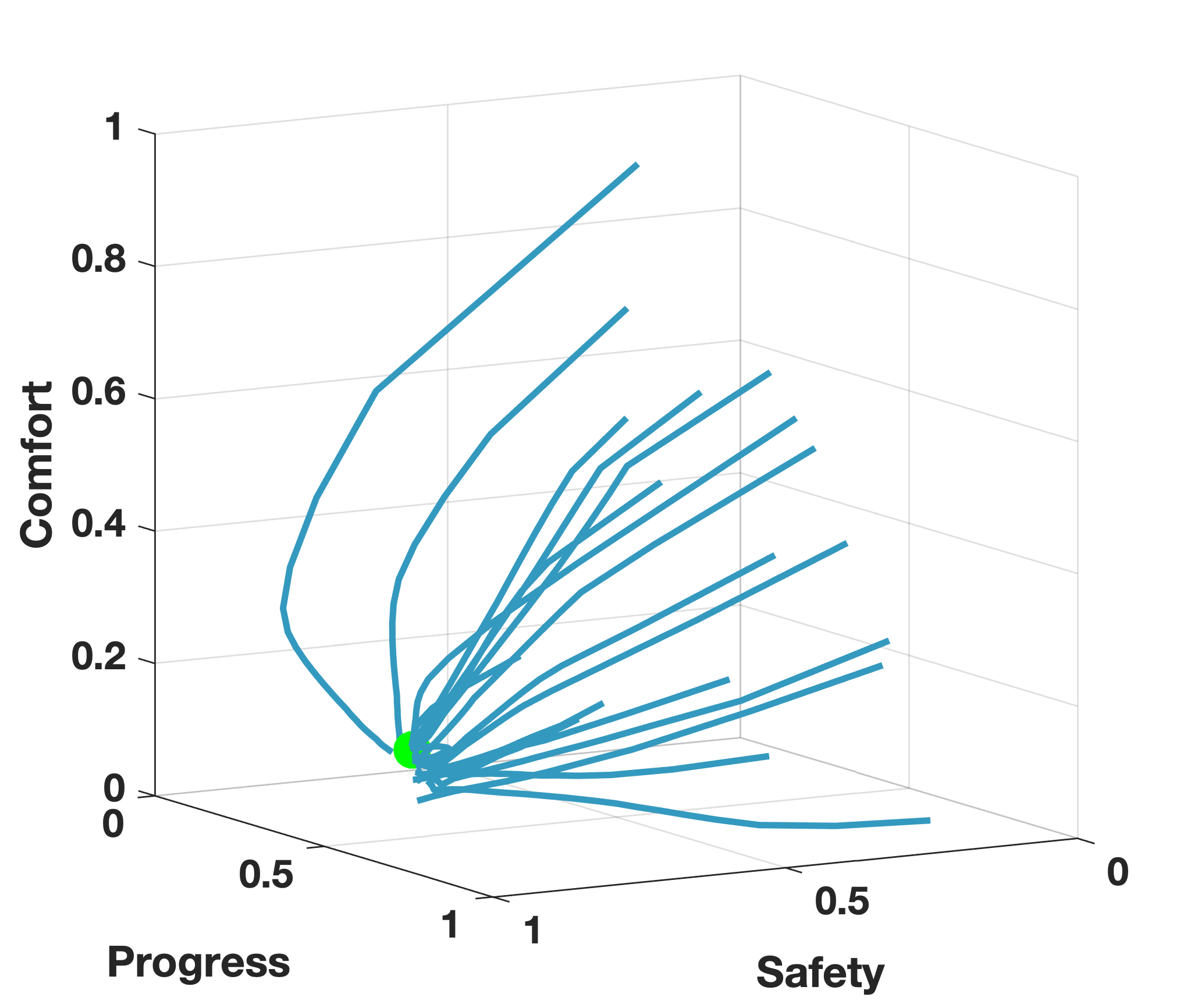, width = 0.33\linewidth, trim=0.2cm 0.0cm 0.0cm 0.5cm,clip}}  %%%
\put(80,  70){\epsfig{file=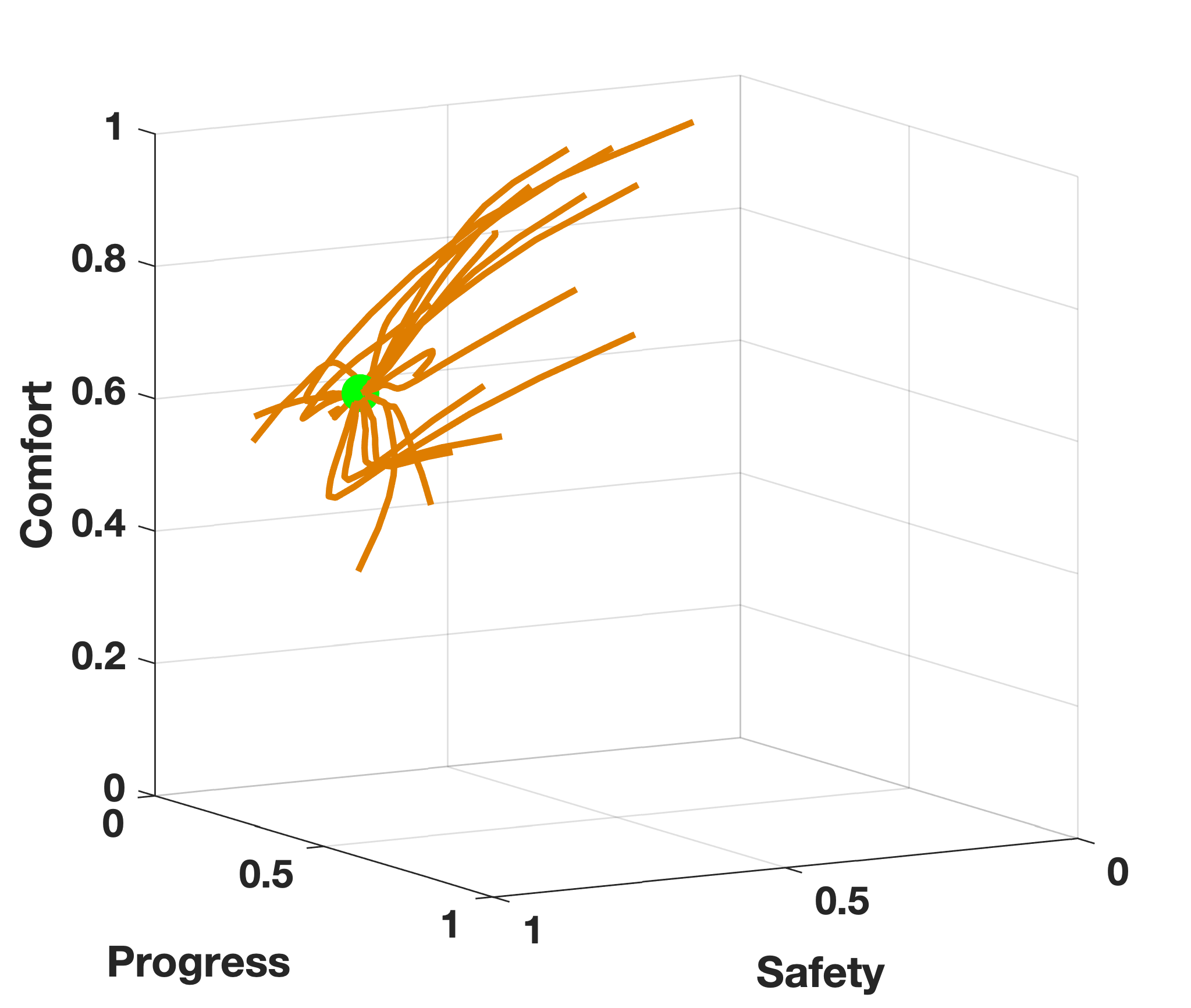, width = 0.33\linewidth, trim=0.2cm 0.0cm 0.0cm 0.5cm,clip}}  %%%
\put(160,  70){\epsfig{file=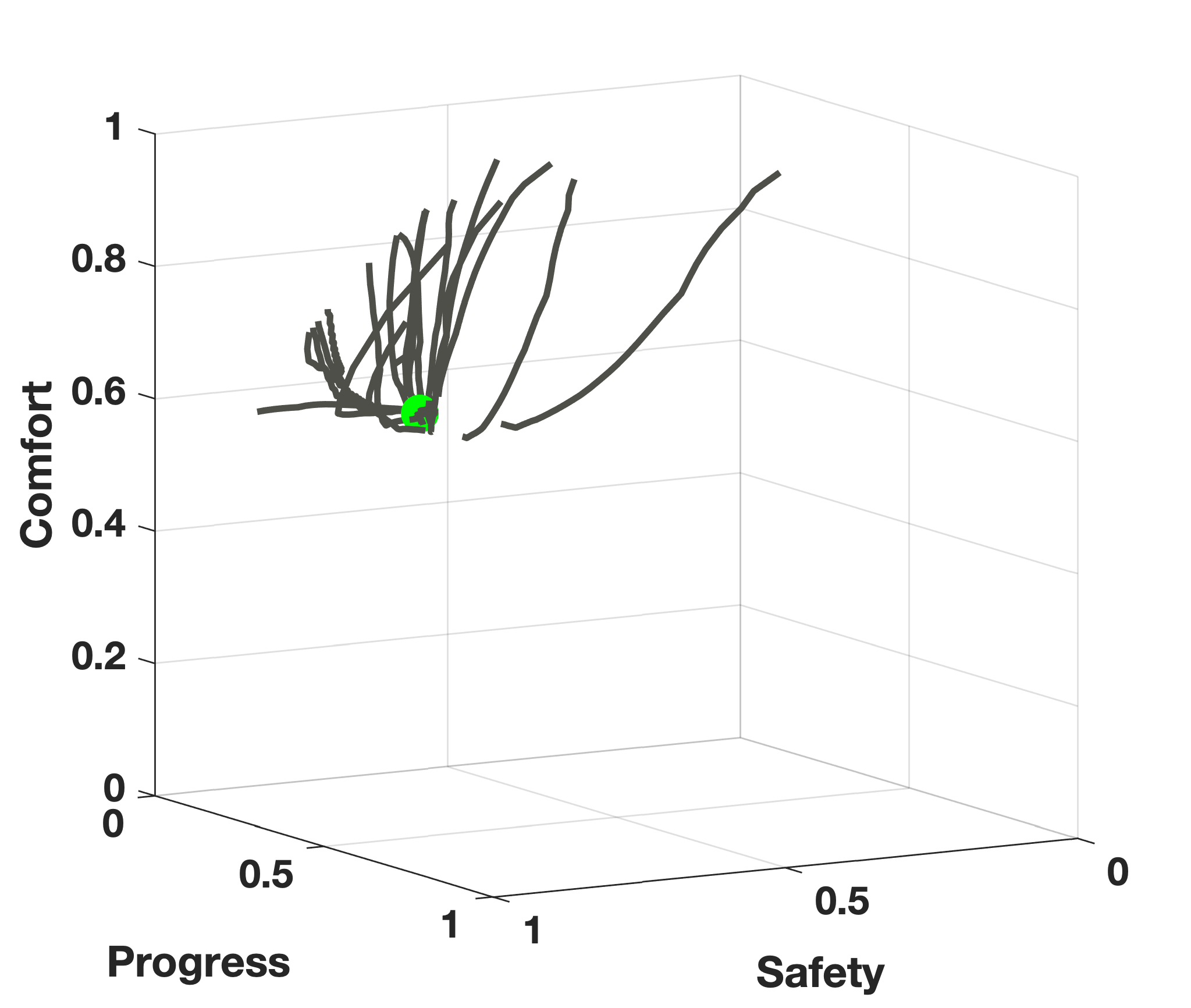, width = 0.33\linewidth, trim=0.2cm 0.0cm 0.0cm 0.5cm,clip}}  %%%
%%%%%%%%%%%%%%%%%%%%%%%%%%%%%%%%%%%%
\put(0,  0){\epsfig{file=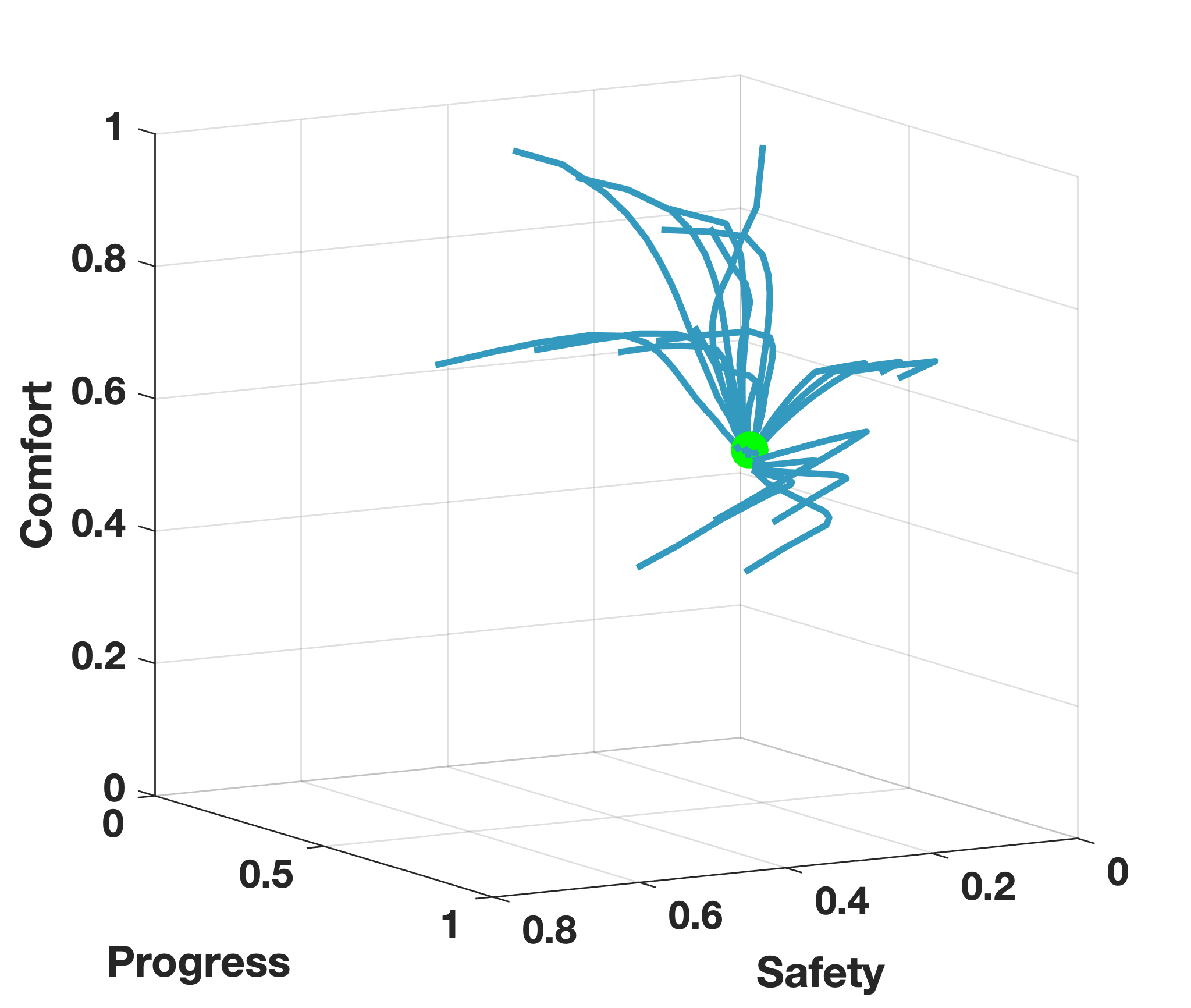, width = 0.33\linewidth, trim=0.2cm 0.0cm 0.0cm 0.5cm,clip}}  %%%
\put(80,  0){\epsfig{file=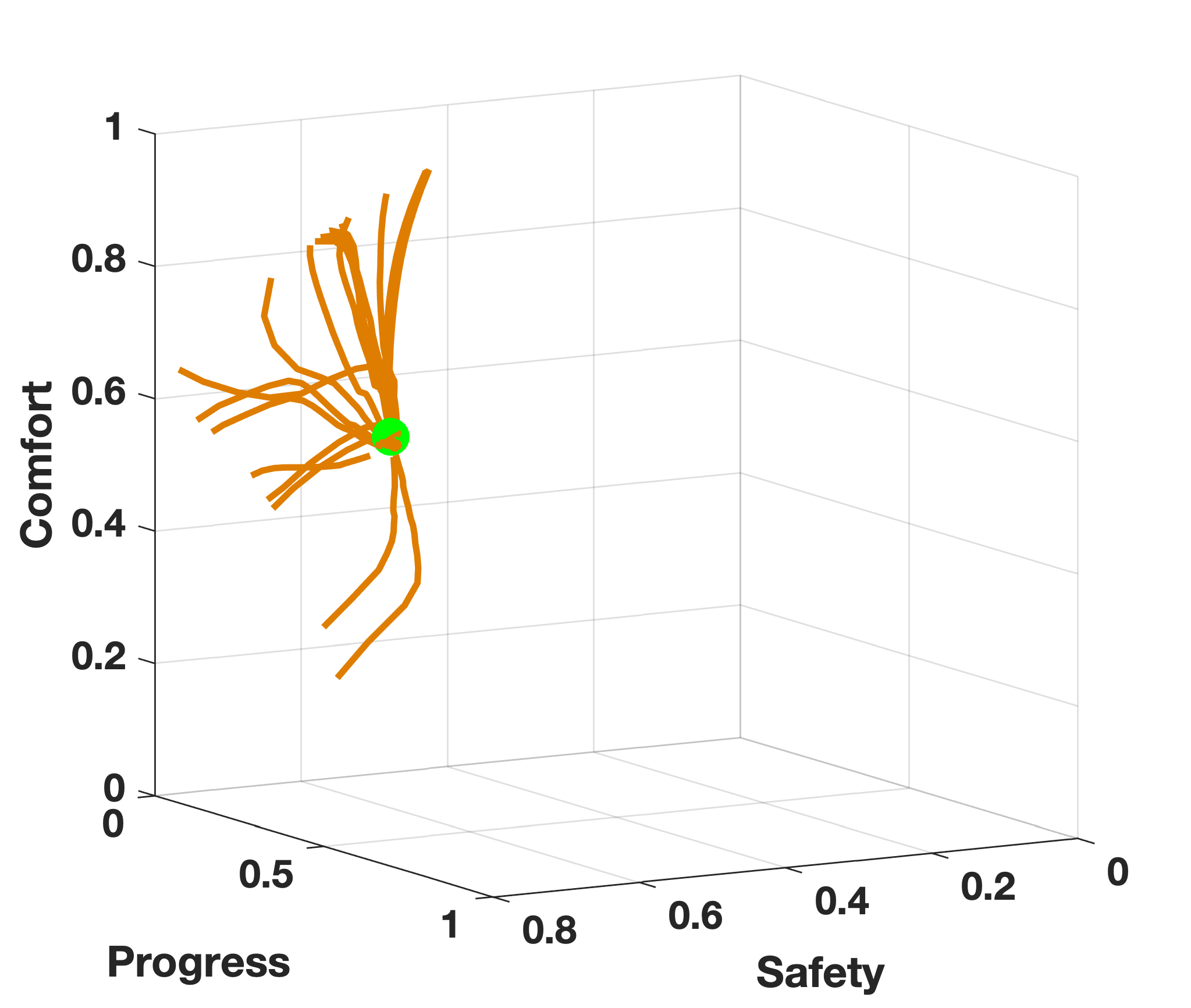, width = 0.33\linewidth, trim=0.2cm 0.0cm 0.0cm 0.5cm,clip}}  %%%
\put(160,  0){\epsfig{file=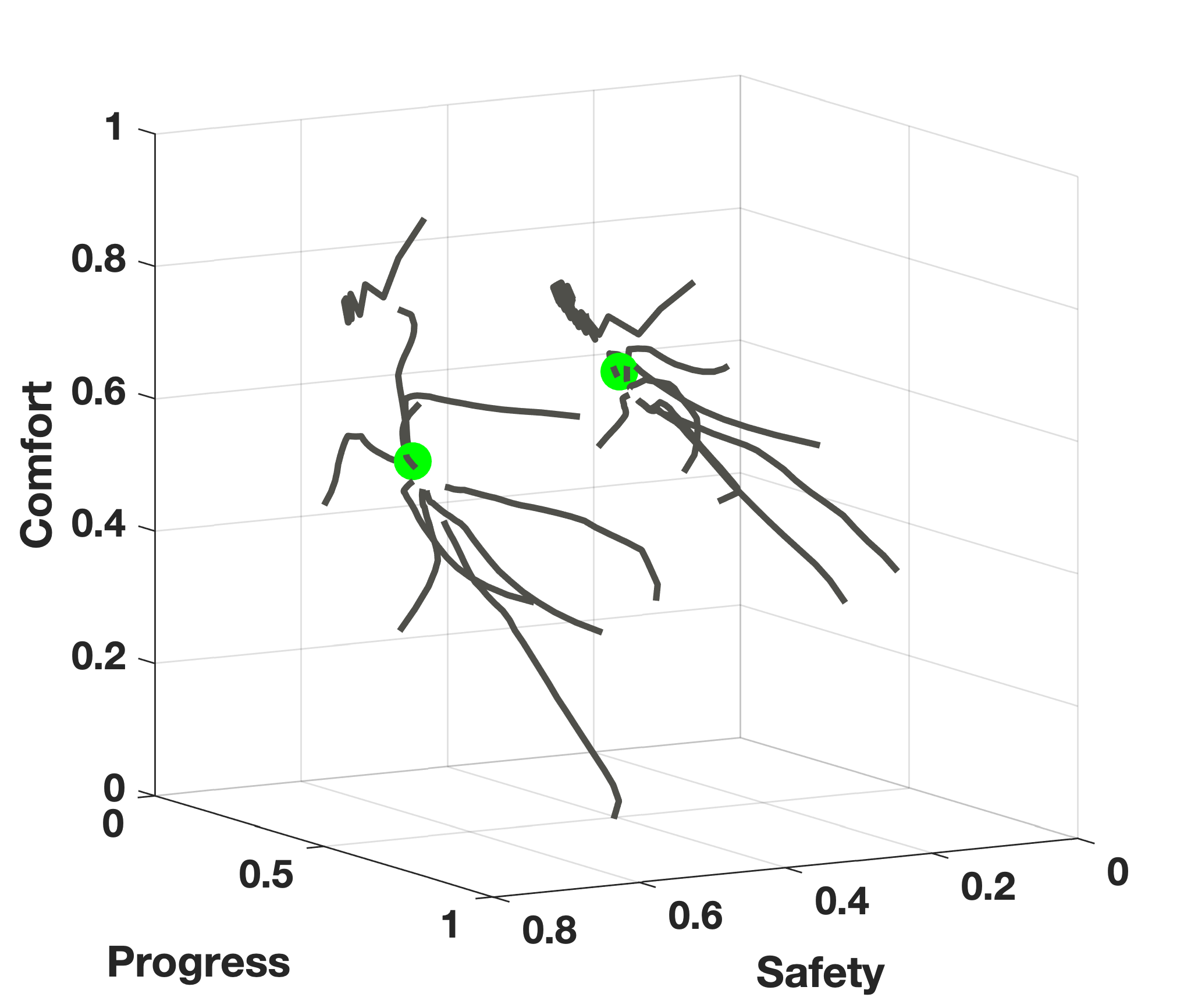, width = 0.33\linewidth, trim=0.2cm 0.0cm 0.0cm 0.5cm,clip}}  %%%
\put(10,120){(a)}
\put(10,50){(b)}
\put(90,120){(c)}
\put(90,50){(d)}
\put(170,120){(e)}
\put(170,50){(f)}
% coordinate
\tiny
% ours
\put(25,80){(0.84,0.53,0.09)}
\put(30,25){(0.33,0.74,0.58)}

% pr2
\put(100,100){(0.76,0.20,0.61)}
\put(100,30){(0.68,0.43,0.60)}

% lf
\put(190,100){(0.73,0.32,0.60)}
\put(180,25){(0.67,0.47,0.56)}
\put(190,55){(0.43,0.57,0.59)}

\end{picture}
\end{center}
\vspace{-0.3cm}
\normalsize
\caption{Histories of feature weights during learning (each line represents a data trail that shows the trajectory of weights during one learning trial). Top row: orange car. Bottom row: blue car. (a-b): Ours. (c-d): PR2-MaxEnt IRL. (e-f): LF-MaxEnt IRL. The green points denote the converged weights in each cluster, and the texts around the green points show the values of the converged weights (safety, progress, comfort). }
\label{fig: weights_hist}
\vspace{-0.4cm}
\end{figure}

\noindent
\textbf{Analysis of the learned driving preferences.} In \cref{fig: weights_hist}, we show the convergence of reward parameters (feature weights) in 15 trials with random initialization. It can be observed that our approach learns that the lower-lane car (orange car) balances progress, comfort, and safety (\cref{fig: weights_hist}(a)), while the upper-lane car (blue car) emphasizes more on progress and comfort (\cref{fig: weights_hist}(b)). Such results align well with our experience and intuition that drivers with right-of-way tend to shift safety responsibilities to others without right-of-way. With PR2-MaxEnt IRL (\cref{fig: weights_hist}(c-d)), due to the homogeneous assumption on the interaction model, the learned weights for both cars are biased towards safety and comfort. With LF-MaxEnt IRL, the weights of the lower-lane car are biased towards safety and comfort (\cref{fig: weights_hist}(e)), and the weights of the upper-lane car converge to two clusters, with one emphasizing more on safety and the other more on progress (\cref{fig: weights_hist}(f)). This is because that LF-MaxEnt IRL pre-assigns the roles of humans, thus it is unable to capture the structural bias in human mind, leading to weights with a large variance.

\begin{table}[ht]
\centering
\caption{Similarity scores of different algorithms}
\vspace{-0.4cm}
\begin{tabular}{c|c c c}
\toprule
\small
Algo. & Ours & PR2-MaxEnt IRL & LF-MaxEnt IRL\\
\hline
Traj. Score & \textbf{10.86} & 17.38 &  24.02\\
\hline
Deci. Score & \textbf{0.93} & 0.73  &  0.6\\
\bottomrule
\end{tabular}
\label{tab: driving-similarity}
\vspace{-0.4cm}
\end{table}
\normalsize

\noindent
\textbf{Analysis of the re-generated interaction trajectories.} Due to the unknown ground-truth reward functions, we investigate whether our approach and the baselines can reproduce interactions in the test set using the learned reward functions and their interaction models. In the right plot of \cref{fig: real_human_loss_hist}, we show an example of the ground-truth interaction trajectory (a) in the test set, and the re-generated interaction trajectories by our approach (b) and the baselines (c-d). The test interaction example is intentionally selected to be complex: both drivers accelerate initially, then the lower-lane driver decelerates slightly but accelerates to merge after observing the deceleration from the upper-lane driver. Our approach is able to generate a seamless interaction that reproduces the behaviors in the test interaction. On the contrary, the baselines are unable to reproduce a similar interaction due to their biased assumptions. Specifically, PR2-MaxEnt IRL assumes that all agents in the game perform L-$1$ recursive reasoning, thus a dead-lock behavior emerged in the generated interaction: both agents tend to yield initially, then the lower-lane driver initiates the merge forced by the approaching dead-end. The leading-following interaction model exploited by LF-MaxEnt IRL assumes one agent aims to compute the best response to the known trajectory of the other agent, thus in the interaction reproduced by LF-MaxEnt IRL, both drivers try to accommodate their opponents' ground-truth trajectories, leading to an incorrect interaction (lower-lane driver yields to the upper-lane driver). In \cref{tab: driving-similarity}, we quantitatively compare the similarity between the re-generated interaction trajectories and the ground-truth ones, at both the trajectory level and the decision level. At the trajectory level, we define similarity score as the (averaged) Euclidean distance between the reproduced trajectories and test trajectories. At the decision level, we define the similarity score as the accuracy of the agents' high-level decisions (yield or not) in the reproduced trajectories with the high-level decisions in the test trajectories as ground-truth decisions. It can be observed from \cref{tab: driving-similarity} that our approach performs significantly better than the baselines.

\vspace{-0.3cm}
